\documentclass{egpubl}
\usepackage{egsgp2024}
\usepackage{amsmath}  
\usepackage{amssymb}
\usepackage{dsfont} 
\usepackage{array}
\usepackage{makecell}
\usepackage{multirow}
\usepackage{color} 
\usepackage{macros}

\SpecialIssuePaper         

\CGFccby

\usepackage[T1]{fontenc}
\usepackage{dfadobe}  

\usepackage{cite}  
\BibtexOrBiblatex
\electronicVersion
\PrintedOrElectronic
\ifpdf \usepackage[pdftex]{graphicx} \pdfcompresslevel=9
\else \usepackage[dvips]{graphicx} \fi
\graphicspath{{./img/}}

\usepackage{egweblnk}

\title[1-Lipschitz Neural Distance Fields]
      {1-Lipschitz Neural Distance Fields}

\author[paper1040]{paper1040}
\author[G. Coiffier \& L. Béthune]{
    \parbox{\textwidth}{
        \centering 
        Guillaume\,Coiffier$^{1}$\orcid{0000-0001-5324-5122}
        and Louis Béthune$^{2,3}$\orcid{0000-0003-1498-8251}
    }
    \\
    \parbox{\textwidth}{
        \centering 
        $^1$Université Catholique de Louvain, Belgium\\
        $^2$IRIT, Université Paul Sabatier, France\\
        $^3$Apple
    }
}

\begin{document}

\teaser{
\centering
\includegraphics[width=\textwidth]{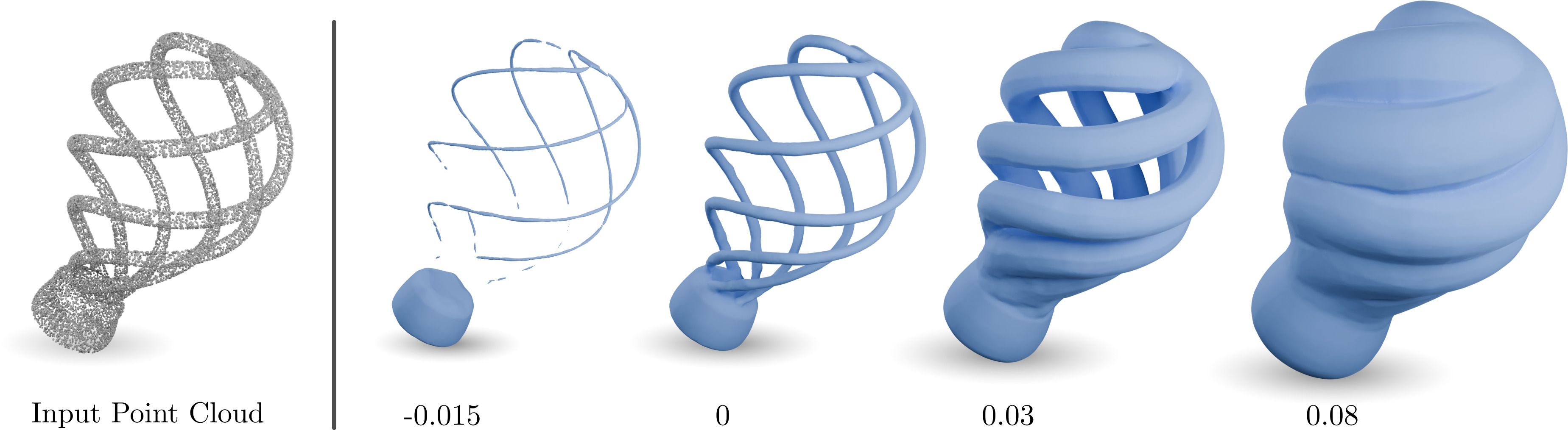} 
\caption{Level sets of a neural distance field trained with our method on a lightbulb model from Thingy10k\cite{Thingi10K}. Given an input point cloud with no knowledge of the ground truth distance function, we are able to fit a neural distance field that is close to the real signed distance function while being guaranteed to be $1$-Lipschitz.}
\label{fig:teaser}
}

\maketitle
\begin{abstract}
   Neural implicit surfaces are a promising tool for geometry processing that represent a solid object as the zero level set of a neural network. Usually trained to approximate a signed distance function of the considered object, these methods exhibit great visual fidelity and quality near the surface, yet their properties tend to degrade with distance, making geometrical queries hard to perform without the help of complex range analysis techniques.
   Based on recent advancements in Lipschitz neural networks, we introduce a new method for approximating the signed distance function of a given object. As our neural function is made $1$-Lipschitz by construction, it cannot overestimate the distance, which guarantees robustness even far from the surface. Moreover, the $1$-Lipschitz constraint allows us to use a different loss function, called the \emph{hinge-Kantorovitch-Rubinstein} loss, which pushes the gradient as close to unit-norm as possible, thus reducing computation costs in iterative queries. As this loss function only needs a rough estimate of occupancy to be optimized, this means that the true distance function need not to be known. We are therefore able to compute neural implicit representations of even bad quality geometry such as noisy point clouds or triangle soups.
   We demonstrate that our methods is able to approximate the distance function of any closed or open surfaces or curves in the plane or in space, while still allowing sphere tracing or closest point projections to be performed robustly.


\end{abstract}


\section{Introduction}

Implicit surfaces~\cite{bloomenthalIntroductionImplicitSurfaces1997} are a powerful tool for geometric modeling and computer graphics, with direct applications in constructive solid geometry, rendering or surface reconstruction. 
Unlike explicit representations like point clouds, surface meshes or voxel grids, which rely on a discretization of space, an implicit representation involves defining an object as the zero level set of a continuous function. In the last few years, this idea have received a lot of attention with the introduction of \emph{neural implicit surfaces}, which encode the function as the parameters of a neural network, allowing such representations to be computed for arbitrary input shapes.

Infinitely many implicit functions can correspond to the same geometry, yet not all of them are created equal: properties of the function sometimes need to also be preserved far from the zero level set. A useful implicit representation to consider in these contexts is
a \emph{signed distance function} (SDF), which outputs the distance to the boundary of its underlying object, counted negatively for points that are inside. A SDF has a unit-norm gradient almost everywhere, making it a $1$-Lipschitz function. Having a $1$-Lipschitz implicit representation is a necessary condition for applications like \emph{ray marching}~\cite{hartSphereTracingGeometric1995}, numerical simulation~\cite{sethianLevelSetMethods2003,sawhneyMonteCarloGeometry2020} or geometrical queries like surface projection to be performed easily. When computing an approximated SDF, it is indeed crucial to never overestimate the true distance otherwise correctness of the queries cannot be guaranteed. In practice, this means that the function's Lipschitz constant should never exceed $1$. However, this Lipschitz property is often overlooked by neural implicit methods, which rather focus on surface fidelity and detail preservation, making them unusable in these contexts without relying on careful range analysis~\cite{sharpSpelunkingDeepGuaranteed2022}.

In this work, we propose a method to approximate the signed distance function of an object by using neural network architectures that are $1$-Lipschitz \emph{by construction}~\cite{araujoUnifiedAlgebraicPerspective2023}, thus guaranteeing correctness of geometrical queries even during training. The Lipschitz constraint of these neural architectures allow us to utilize a different loss function, called the \emph{hinge-Kantorovitch-Rubinstein} (hKR) loss~\cite{serrurierAchievingRobustnessClassification2021}. This loss has two important effects. Firstly, we prove that any minimizer over all possible Lipschitz functions is close to the SDF of the considered object, which makes our trained neural network a very good approximation of the true distance, as illustrated in Figure~\ref{fig:teaser}. Secondly, using the hKR loss makes us approach the problem of learning a signed distance field not from the usual \emph{supervised} regression point of view but from a \emph{semi-supervised} classification point of view: instead of fitting a neural network's output to precomputed distances over a dataset of points, we instead try to maximize the distance between points from inside the shape and points from outside while remaining $1$-Lipschitz. This means in particular that the only information required for training is knowing in which category (inside or outside of the input shape) a point is, an information that can be robustly extracted even for point clouds or triangle soups~\cite{barillFastWindingNumbers2018}. As a consequence, we are able to approximate the SDF of an object \emph{without access to the ground truth distance}, enabling training for a wide range of inputs including triangle soups and point clouds, even noisy, sparse or incomplete.

To summarize, our contributions are as follows:
\begin{itemize}
    \item We apply the known method of minimizing the hKR loss on some $1$-Lipschitz neural network to the problem of approximating the signed distance field of an object.
    \item We demonstrate that such an approach solves the usual robustness issues of similar methods, as it outputs a function that is a good approximation of the true signed distance function while being guaranteed to never overestimate it.
    \item As the hKR loss does not need ground truth distances but only occupancy labels, we show that we are able to compute signed or unsigned distance fields of noisy, incomplete or sparse representations of objects of any topology, including open surfaces and curves.
    \item We apply our method to a variety of geometry processing tasks, like surface sampling, medial axis estimation, constructive solid geometry and ray marching.
\end{itemize}
\section{Background and Related Work}

\subsection{Signed Distance Function}
\label{ssec:SDF}
In all of this work, we will denote by $\Omega$ some solid object $\real^n$, where $n=2$ or $3$. The \emph{signed distance function} (SDF) of $\Omega$ is the function $S_\Omega$ defined over $\real^n$ as:



$$ S_\Omega(x) = \left(\mathds{1}_{\real^n \backslash \Omega}(x) - \mathds{1}_{\Omega}(x) \right) \min_{p \in \partial \Omega} ||x - p||$$

where $\partial \Omega$ is the boundary of $\Omega$ and the distance considered in the Euclidean distance.

\begin{figure}[b]
    \centering
    \def\svgwidth{\linewidth}
\begingroup%
  \makeatletter%
  \providecommand\color[2][]{%
    \errmessage{(Inkscape) Color is used for the text in Inkscape, but the package 'color.sty' is not loaded}%
    \renewcommand\color[2][]{}%
  }%
  \providecommand\transparent[1]{%
    \errmessage{(Inkscape) Transparency is used (non-zero) for the text in Inkscape, but the package 'transparent.sty' is not loaded}%
    \renewcommand\transparent[1]{}%
  }%
  \providecommand\rotatebox[2]{#2}%
  \newcommand*\fsize{\dimexpr\f@size pt\relax}%
  \newcommand*\lineheight[1]{\fontsize{\fsize}{#1\fsize}\selectfont}%
  \ifx\svgwidth\undefined%
    \setlength{\unitlength}{884.98419814bp}%
    \ifx\svgscale\undefined%
      \relax%
    \else%
      \setlength{\unitlength}{\unitlength * \real{\svgscale}}%
    \fi%
  \else%
    \setlength{\unitlength}{\svgwidth}%
  \fi%
  \global\let\svgwidth\undefined%
  \global\let\svgscale\undefined%
  \makeatother%
  \begin{picture}(1,0.36379034)%
    \lineheight{1}%
    \setlength\tabcolsep{0pt}%
    \put(0,0){\includegraphics[width=\unitlength,page=1]{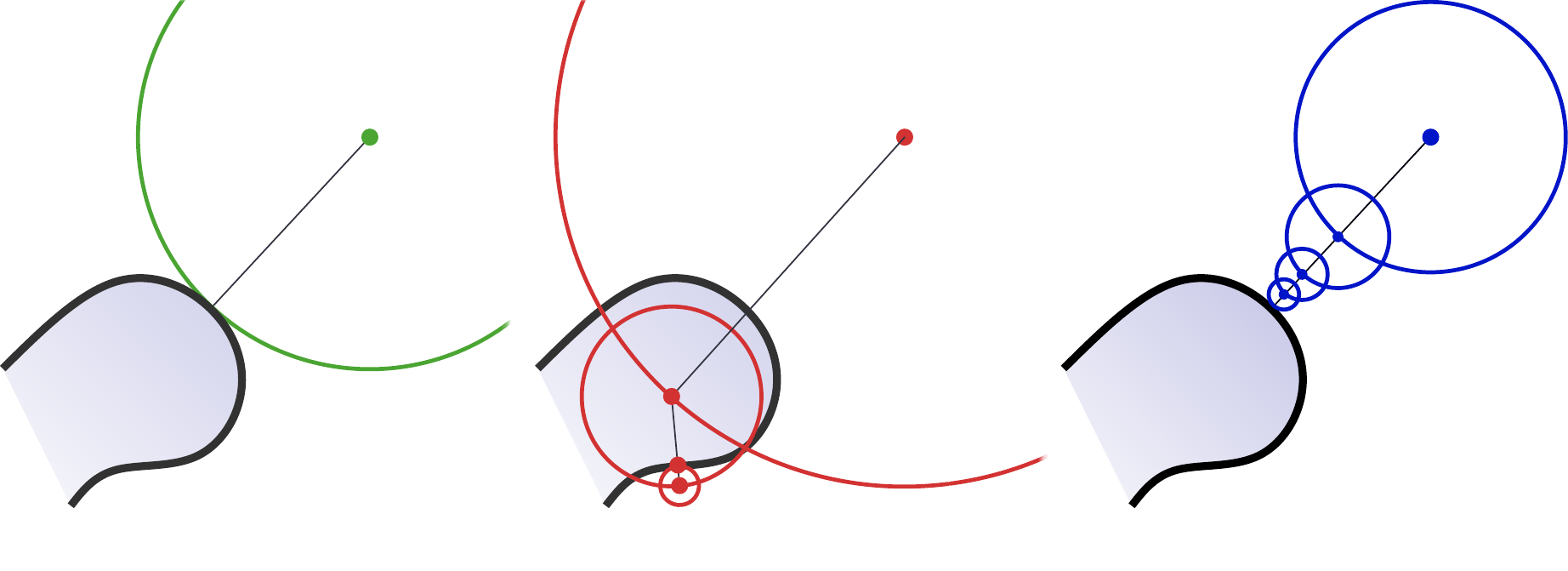}}%
    \put(0.68422524,0.0063891){\color[rgb]{0,0,0}\makebox(0,0)[lt]{\lineheight{1.25}\smash{\begin{tabular}[t]{l}$|f|<|S_\Omega|$\end{tabular}}}}%
    \put(0.38394475,0.0063891){\color[rgb]{0.02745098,0.02745098,0.02745098}\makebox(0,0)[lt]{\lineheight{1.25}\smash{\begin{tabular}[t]{l}$|f|>|S_\Omega|$\end{tabular}}}}%
    \put(0.07692348,0.00650816){\color[rgb]{0,0,0}\makebox(0,0)[lt]{\lineheight{1.25}\smash{\begin{tabular}[t]{l}$S_\Omega$\end{tabular}}}}%
  \end{picture}%
\endgroup%

    \caption{The value of the SDF $S_\Omega$ at point $x$ is the radius of the larger sphere centered at $x$ that does not intersect $\Omega$ (left). For an approximated SDF $f$, if $f$ overestimates the distance (middle), then the query is wrong and no guarantees can be drawn. If $f$ always underestimates the distance (right), iterating the query still converges to the correct result.}
    \label{fig:SDF_projection}
\end{figure}

Signed distance functions have mainly been studied in computer graphics for the ease with which they enable certain operations like boolean composition~\cite{ricciConstructiveGeometryComputer1973}, smooth blending~\cite{blinnGeneralizationAlgebraicSurface1982}, surface offset~\cite{friskenDesigningDistanceFields2006} or deformation~\cite{sederbergFreeformDeformationSolid1986a} while still allowing an explicit representation, like a surface mesh, to be extracted for instance using the \emph{marching cubes} algorithm~\cite{lorensenMarchingCubesHigh1987,dearaujoSurveyImplicitSurface2015}. In essence, the SDF value at point $x$ gives two pieces of information: its sign directly tells if the query point is inside or outside the object, while its magnitude gives the radius of the largest sphere centered at $x$ that does not intersect the boundary of the object. This observation is the starting point of the \emph{sphere tracing} algorithm~\cite{hartSphereTracingGeometric1995} which enables direct rendering of SDFs. Additionally, the gradient of the SDF is aligned with the normal vector field of the object on its boundary and gives the direction to the closest point on the boundary. Evaluating the function and its gradient at a point therefore gives a simple strategy for projecting onto the zero level set (Figure~\ref{fig:SDF_projection}, left).
 




\subsection{Lipschitz Implicit Representations}
\label{ssec:bg_queries}

Although the SDF of an object is easy to compute in closed form for simple shapes (see for instance~\cite{QuilezSDF} for a list), the same cannot be said of objects found in the wild. Representing those objects via a general implicit surface or even an approximated SDF can still achieve high visual fidelity but comes at a cost. In such contexts, there is indeed no direct strategy for closest point queries or ray intersections: one has to rely to iterative methods, where a key quantity to control is the Lipschitz constant $L$ of the implicit function. Recall that a function $f$ is $L$-Lipschitz if it satisfies:

$$\forall a,b \in \real^n, \quad ||f(b) - f(a)|| \leqslant L\,||b-a||.$$

If $f$ is differentiable, its Lipschitz constant $L$ is an upper bound on the norm of its gradient. Since the signed distance function has a unit-norm gradient, an implicit function with $L>1$ may overestimate the true distance. As a consequence, the sphere centered at $x$ of radius $f(x)$ may intersect the surface and yield a false negative (Figure~\ref{fig:SDF_projection}, middle). No guarantee can be drawn onto the correctness of the query in this case. On the other hand, having $L<1$ implies that the function always underestimates the distance. Iterating the empty sphere query in this case will converge to the correct result at a cost of a greater computation time (Figure~\ref{fig:SDF_projection}, right).

An extensive bibliography exists on algorithms for performing robust geometric queries on approximate SDFs. Previous works can be organized into two categories. Methods from the first one rely on interval arithmetic and careful range analysis to detect overshooting and false negatives~\cite{duffIntervalArithmeticRecursive1992,sharpSpelunkingDeepGuaranteed2022,aydinlilarForwardInclusionFunctions2023}, thus guaranteeing robustness at the cost of more complexity. Methods from the second category instead estimate the Lipschitz constant of the implicit function and apply a local or global rescaling~\cite{kalraGuaranteedRayIntersections1989, galinSegmentTracingUsing2020} to query $f/L$ instead. However, computing the exact Lipschitz function for neural networks is a NP-hard problem~\cite{jordanExactlyComputingLocal2020} and finding a good approximation of it remains tricky~\cite{virmauxLipschitzRegularityDeep2018}.

\subsection{Neural Implicit Surfaces}
\label{ssec:NDF}

Approximating a distance function has historically been achieved using basis functions like blobs or blended balls~\cite{blinnGeneralizationAlgebraicSurface1982,wyvillDataStructureForsoft1986}. In the last few years, an ongoing trend has proposed to encode it into the parameters $\theta$ of a multilayer perceptron (MLP) $f_\theta$. This idea of a neural field has sparked many applications in computer graphics and learning, which are not restricted to distance fields. We refer to Xie et al.~\cite{xieNeuralFieldsVisual2022} for a survey.

Perhaps the most simple neural implicit representation is to represent the shape as a binary occupancy field, predicting $1$ for points inside the shape and $0$ otherwise~\cite{chenLearningImplicitFields2019}. This can be seen as a binary classification problem and treated as such~\cite{meschederOccupancyNetworksLearning2019}. While enabling total surface reconstruction, such neural fields give no information far from the surface and are therefore hard to query geometrically.

\begin{figure}
    \centering
    \def\svgwidth{\linewidth}
    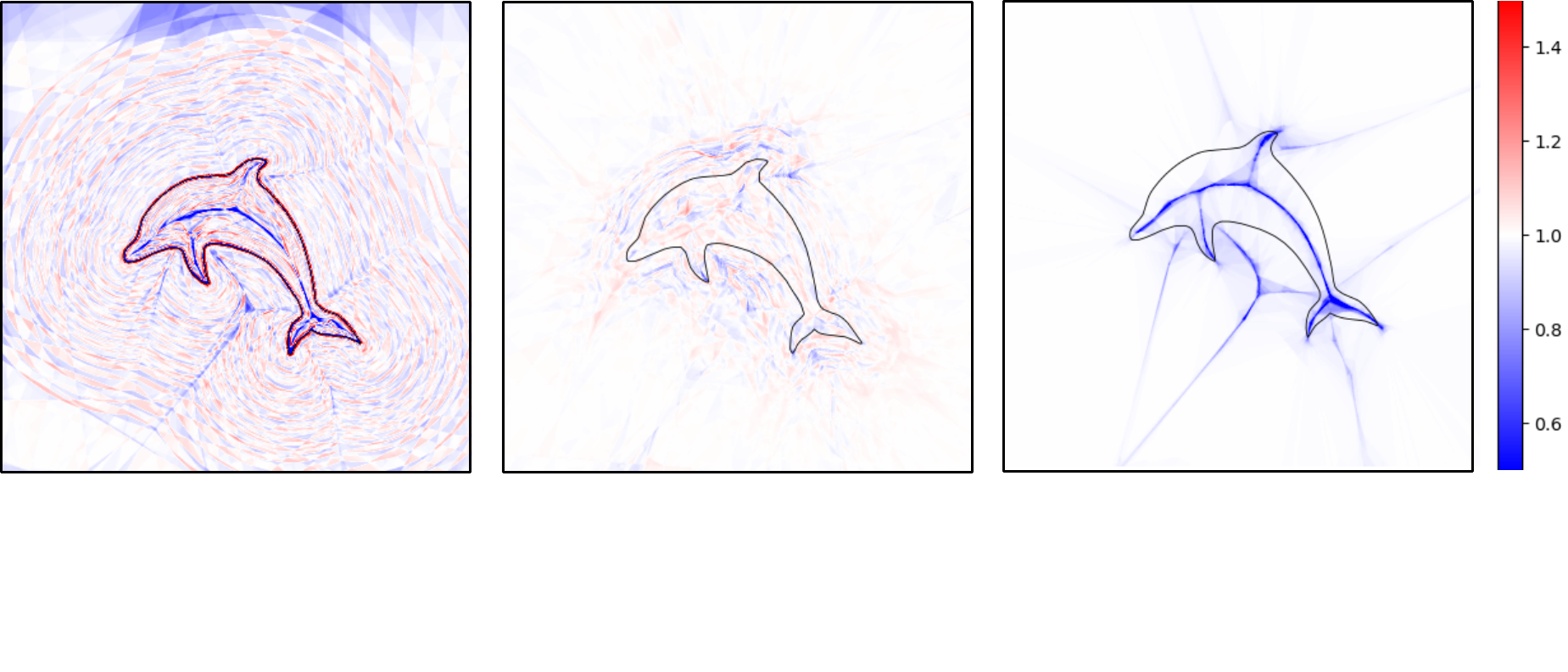
    \caption{Plot of a the gradient norm of neural distance fields on a simple 2D dolphin silhouette. While minimizing the eikonal loss stabilizes the gradient norm, only a Lipschitz network guarantees a unit bound.}
    \label{fig:eikonal_cest_dla_merde}
\end{figure}

The \emph{DeepSDF}~\cite{parkDeepSDFLearningContinuous2019} algorithm is the first proposition of a neural SDF. It is setup as a single large neural network optimized over a collection of objects, where a given object is represented via a latent vector fed as an input along the query point. This popular setup allows shape interpolation~\cite{liuLearningSmoothNeural2022}, classification as well as shape segmentation~\cite{petrovANISEAssemblybasedNeural2023}. 
In contrast, training one network per object has also been performed~\cite{daviesEffectivenessWeightEncodedNeural2021} for shape compression purposes. Learning an \emph{unsigned} distance field has also been attempted either directly~\cite{chibaneNeuralUnsignedDistance2020} or using a sign-agnostic loss~\cite{atzmonSALSignAgnostic2020a}, thus extending the application of neural distance fields to open surfaces and curves. All of these methods are \emph{supervised}, meaning that they are optimized to make the network's predictions match some pre-computed signed distances. Unfortunately, the minimization of this loss alone does not guarantee that the network will correctly extrapolate the distance for points not in the dataset, especially if it is not dense enough. In particular, the resulting function may present a gradient whose norm may vary extensively and thus a large Lipschitz constant. This phenomenon is illustrated on a simple 2D dataset on Figure~\ref{fig:eikonal_cest_dla_merde} (a).

In order to bring the Lipschitz constant closer to $1$, many works rely on regularization losses applied to the gradient of the network. The most widespread of such regularization is the \emph{eikonal} loss, which computes how much the norm of the gradient differs from $1$. Initially introduced to improve the stability of the \emph{Wasserstein Generative Adversarial Networks}~\cite{gulrajaniImprovedTrainingWasserstein2017}, it was then naturally adapted to neural distance fields~\cite{groppImplicitGeometricRegularization2020, yarivVolumeRenderingNeural2021}. An alternative to the eikonal loss is the total variation loss~\cite{clemotNeuralSkeletonImplicit2023}, which minimizes variations of the gradient norm and improves the quality of the neural implicit function far from the zero level set.
Alignment losses, that penalize the difference between the gradient and some normal vector field, improving visual fidelity over the zero level set, have also been considered~\cite{atzmonSALDSignAgnostic2020a, sitzmannImplicitNeuralRepresentations2020}, often alongside some eikonal term.

Yet, while these regularizing losses have a clear impact on the gradient of the considered implicit function, their minimization is performed in practice using first-order optimizers like \emph{Adam}~\cite{kingma2014adam}, which means that their global minimizer is never reached. Even though specific neural architectures have been designed to better behave during optimization, like \emph{SIREN}~\cite{sitzmannImplicitNeuralRepresentations2020} that uses sine activations, nothing prevents a regularized network to have a Lipschitz constant larger than $1$ even after training for a long period of time, as shown in Figure~\ref{fig:eikonal_cest_dla_merde} (b). 

Few neural implicit methods tackle the case where the true signed distance to the object is not known or cannot be computed. In this harder context, Lipman~\cite{lipmanPhaseTransitionsDistance2021a} defines the \emph{PHASE} loss, allowing to learn an occupancy field using only a representation of the boundary of the object. The signed distance function is then retrieved using a log transformation. Finally, while their application is classification and outlier detection, the method of Béthune et al.~\cite{bethuneRobustOneClassClassification2023} learns a SDF from an occupancy field by also minimizing the \emph{hinge-Kantorovitch-Rubinstein} loss. Although very similar, our method is simpler as their Newton-Raphson update of the distribution is unnecessary in our case, as well as faster since we use a more computationally efficient neural architecture.

\begin{figure*}
    \centering
    \includegraphics[width=\textwidth]{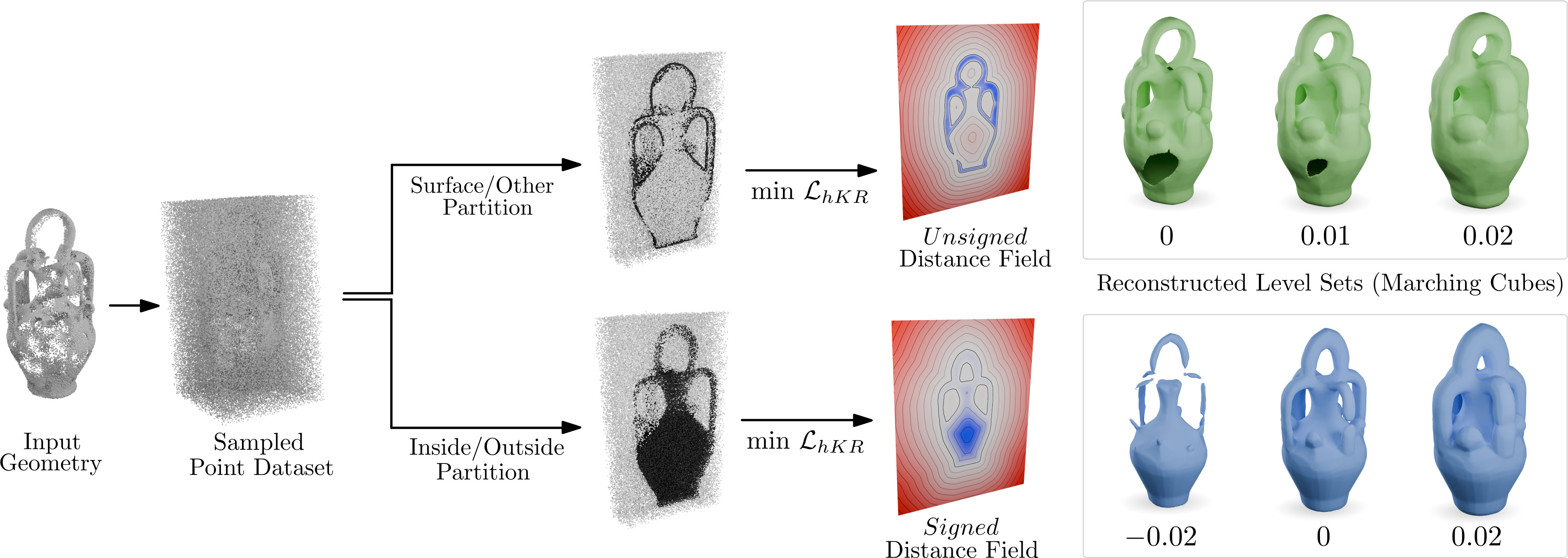}
    \caption{Overview of our method on a corrupted \emph{Botijo} dataset. Given an input geometry in the form of an oriented point cloud or a triangle soup, we uniformly sample points in a domain containing the desired geometry. Defining negative samples as points of the geometry and positive samples everywhere else yields an unsigned distance field when minimizing the hKR loss. On the other hand, partitioning samples as \textit{inside} or \textit{outside} the shape leads to an approximation of the signed distance function of the object.}
    \label{fig:overview}
\end{figure*}

\subsection{Lipschitz Neural Networks}
\label{ssec:lip_net}

At its core, a neural network is nothing more than a function $f_\theta$ where the parameters (or weights) $\theta \in \real^K$ are arranged in a predetermined pattern. Specifying such an architecture defines a functional space:
$$\mathcal{F} = \{ f_\theta \,|\, \theta \in \real^K \}$$
over which learning algorithms optimize the weights to find the function $f_\theta^*$ that minimizes some user-defined loss criterion. Knowing exactly the extent of the functional space $\mathcal{F}$ for a fixed architecture is an open problem in deep learning, but recent works have managed to define some $\mathcal{F}$ as a subset of $L$-Lipschitz functions~\cite{serrurierAchievingRobustnessClassification2021}.

The investigation of Lipschitz architecture in deep learning is primarily motivated by the robustness of such networks against adversarial attacks and overfitting. Early attempts focused on regularizing the weight matrices by controlling their largest singular value~\cite{yoshidaSpectralNormRegularization2017}. To prevent the gradient from vanishing, other efforts focused on having singular values all close to one, namely regularizing weight matrices to be orthogonal~\cite{cisseParsevalNetworksImproving2017,trockmanOrthogonalizingConvolutionalLayers2020} and the network to be gradient preserving. Since this hindered expressiveness overall when using usual component-wise activation functions like the sigmoid or ReLU, Anil et al.~\cite{anilSortingOutLipschitz2019} propose a sort as a non-linearity. Lispchitz network have since been shown to have comparable results with classical neural networks on a variety of tasks~\cite{bethunePayAttentionYour2022}.

More recently, other constructions of Lipschitz neural layers decreasing the computational cost of previous approaches have been proposed. Instead of relying on iterative projections of weight matrices to orthogonal ones, Prach and Lampert~\cite{prachAlmostOrthogonalLayersEfficient2022} introduce an almost orthogonal layer where singular values of the matrices are updated directly during training. Other Lipschitz layers have then been defined, such as the \emph{Convex Potential Layer}~\cite{meunierDynamicalSystemPerspective2022} or the \emph{Semi-definite Programming Lipschitz Layer} (SLL)~\cite{araujoUnifiedAlgebraicPerspective2023}. Neural architectures used in this work are based on the latter.

\section{Robust Learning of a Signed Distance Function}

As current neural implicit representations cannot provide guarantees on geometrical queries from the implicit function nor theoretical bounds on its Lipschitz constant, we propose to directly integrate the constraint of being $1$-Lipschitz directly into the neural architecture. As shown experimentally in Figure~\ref{fig:eikonal_cest_dla_merde} (c), this will result in an implicit function that cannot overestimate the true distance by construction, even during training. 

Our method takes as an input any curve or surface $\partial \Omega$ from $\real^n$, represented either by a point cloud with normals or a triangle soup. The first step is to define some $1$-Lipschitz neural architecture, for which we use the \emph{SLL} architecture of Araujo et al.~\cite{araujoUnifiedAlgebraicPerspective2023} (Section~\ref{ssec:lip_archi}). As having a Lipschitz constant strictly smaller than $1$ can induce greater computation times for geometrical queries to converge, it is desired to not only be $1$-Lipschitz but to have gradient as close as possible to unit norm everywhere. In our case, this is achieved by the \emph{hinge-Kantorovitch-Rubinstein} (hKR) loss whose indirect effect is to maximize the gradient's norm (Section~\ref{ssec:hkr}). Minimizing the hKR loss requires to partition a dataset of points around the object as points inside and outside. In the case of closed surfaces or curves, we use the \emph{generalized winding number}~\cite{barillFastWindingNumbers2018} to robustly compute this partition (Section~\ref{ssec:winding}). Finally, the case of open surfaces and curves, for which we want to compute an unsigned distance function, will be discussed in Section~\ref{ssec:no_interior}. The different steps of our method are illustrated on Figure~\ref{fig:overview} on a corrupted point cloud of the \emph{Botijo} model.

\subsection{$1$-Lipschitz Neural Architecture}
\label{ssec:lip_archi}

Classically, a neural network $f_\theta$ has its parameters $\theta$ arranged in a series of layers $f^1,...,f^l$ so that the final function is the composition of all layers in order. As the Lipschitz constant of a composition is upper bounded by the product of all Lipschitz constants, designing a $1$-Lipschitz architectures boils down to defining some $1$-Lipschitz layers to be chained together. To this end, Araujo et al.~\cite{araujoUnifiedAlgebraicPerspective2023} propose the \emph{Semi-definite Programming Lipschitz Layer} (SLL). Using a square matrix $W \in \real^{k \times k}$, a bias vector $b \in \real^k$ and an additional vector $q \in \real^k$ as parameters, it is defined as:

\begin{equation}
    x \mapsto x - 2WT^{-1} \sigma(W^Tx + b)
    \label{eq:SLL}
\end{equation}
where $T$ is a diagonal matrix of size $\real^{k \times k}$:
$$ T_{ii} = \sum_{j=1}^k \left| (W^T W)_{ij} \,\exp(q_j - q_i) \right|$$

and $\sigma(x) = \max(0, x)$ is the rectified linear unit (ReLU) function. In comparison to the classical multilayer perceptron layer $x \mapsto \sigma(W^Tx +b)$, the SLL layer only adds a small amount of parameters in the form of the vector $q$ and a $\mathcal{O}(k^2)$ operations, which makes it more efficient than previous Lipschitz architectures. 

The SLL function is a residual layer, meaning that its computation is added to its input. As a consequence, the layer can only be defined for matching input and output dimensions. In our case, the input of the network is a point in $\real^n$ with $n=2$ or $3$ and its output is a single real number. The input of the network is therefore first padded with zeros to match the size $k$ of the SLL layers. To retrieve a single number as output, the network ends with an affine layer defined as:
$$ x \mapsto \frac{w^Tx}{||w||_2} + b$$
where $w \in \real^k$ and $b \in \real$. Dividing by the euclidean norm of $w$ in the computation ensures that the operation is $1$-Lipschitz.

\subsection{The hinge-Kantorovitch-Rubinstein Loss Function}
\label{ssec:hkr}

Given a dataset $(X,Y)$ of points with associated signed distance ground truth, a straightforward approach to learn a neural signed distance field guaranteed to always underestimate the true distance would be to minimize a fitting loss over a $1$-Lipschitz architecture as defined above. While it solves the problem of robustness of neural signed distance fields, this approach fails short of our goal for two reasons. Firstly, this means that the input geometry must be well known for the initial dataset to be computed with exact signed distances, making it unsuitable for only raw point clouds or triangle soups as inputs. Secondly, the function will indeed be $1$-Lipschitz but can greatly underestimate the true distance. Ideally, we would like to also maximize its gradient norm so that fewer iterations are needed in geometrical queries.

These two limitations can be overcome with the \emph{hinge-Kantorovitch-Rubinstein} (hKR) loss. Introduced by Serrurier et al.~\cite{serrurierAchievingRobustnessClassification2021}, the hKR is a binary classification loss that can be optimized over some $L$-Lipschitz architecture. Let $D \subset \real^n$ be a compact domain over which binary labels $y(x) \in \{-1,1\}$ are defined at each points. Let $\lambda>0$ and $m>0$ be hyperparameters. Let $\rho(x)$ be some probability distribution function over $D$, which can be thought of as an importance weight. The hKR loss of a neural function $f_\theta$ is the sum of two terms:
\begin{equation}
    \loss_{hKR} = \loss_{KR} + \lambda \loss_{hinge}^m
    \label{eq:hkr}
\end{equation}
defined as:
\begin{align}
\loss_{KR}(f_\theta,y) &= \int_D -y(x) f_\theta(x) \, \rho(x)dx \label{eq:hkr1} \\
\loss_{hinge}^m(f_\theta,y) &= \int_D \max\left(0, m-y(x)f_\theta(x)\right)\,\rho(x)dx \label{eq:hkr2}.
\end{align}

The first term (Equation~\eqref{eq:hkr1}), minimized over all possible $1$-Lipschitz functions, is known in the literature as the dual formulation of the Wasserstein-1 distance~\cite{arjovskyWassersteinGAN2017}, as given by the Kantorovitch-Rubinstein duality theorem~\cite[Theorem 5.10]{villani2009optimal}. While its connections to optimal transport are out of scope of this work, it can be interpreted as maximizing values of $f_\theta$ for points where $y=1$ and minimizing them when $y=-1$. It therefore encourages the function to maximize its rate of change, and thus its Lipschitz constant. However, as observed by Serrurier et al.~\cite{serrurierAchievingRobustnessClassification2021}, simply optimizing $\loss_{KR}$ leads to bad results as the zero level set of the network does not capture the boundary between positive and negative $y$. This is where the second term comes into play: the hinge loss of Equation~\eqref{eq:hkr2} penalizes points for which the sign of $f_\theta$ and $y$ are different, that is to say points that are "misclassified" by $f_\theta$. The parameter $m>0$, called \emph{margin}, defines an error threshold under which this misclassification is ignored. Low values of $m$ lead to more precise results at the interface but also unstability in the optimization.

In the case of the neural distance field of an object $\Omega$, we can apply the hKR loss by considering $y$ as an occupancy label, being $-1$ for points inside of $\Omega$ and $+1$ otherwise. In this context, under mild assumptions over $\rho$, minimizers of the hKR loss are good approximations of the signed distance function of $\Omega$. This is summarized as the following theorem:

\begin{theorem}
    \itshape Let $D,\Omega$ be compact subsets of $\real^n$ such that $\Omega \subset D$. Let $y$ be binary labels defined over $D$ as: 
    $$y(x) = \mathds{1}_{D \backslash \Omega}(x) - \mathds{1}_{\Omega}(x).$$
    Let $m>0$ and assume that $\rho(x)=0$ whenever $|S_\Omega(x)|\leqslant m$ and $\rho(x)>0$ otherwise. 
    
    Let $f^*$ be a minimizer of $\loss_{KR}(f,y)$ under constraint that $\loss_{hinge}^m(f,y) = 0$, where the minimum is taken over all possible $1$-Lipschitz functions. Then:

    $$\forall x \in D, \quad \left\{\begin{array}{rcl}
        |S_\Omega(x)| > m & \implies & f^*(x) = S_\Omega(x) \\
        |S_\Omega(x)| \leqslant m &\implies & |f^*(x) - S_\Omega(x)| \leqslant 2m\\
    \end{array}\right..$$
\label{thm:THE_THEOREM}
\end{theorem}
This theorem is very similar to the result of Béthune et al.~\cite[Theorem~1]{bethuneRobustOneClassClassification2023}, which shows a similar result in a context where $\rho(x)$ is zero whenever $0 \leqslant S_\Omega(x) \leqslant 2m$. This translated version is relevant to their application of outlier detection; in our case however, we focus on the "symmetric" version. A fully detailed proof is available in Appendix~\ref{sec:proof_of_thm}.

The key idea of the proof is to split the domain $D$ into two parts: the region where $\rho = 0$, which corresponds to a "shell" set  of width $2m$ centered on $\partial \Omega$, and its complementary. This allows to exploit the fact that $\loss_{hinge}^m(f^*,y) = 0$ and deduce the approximation by at most $2m$ in this region. Outside of this region, one can show that $f^*$ has the same sign as $S_\Omega$, and its Lipschitz property means that it is bounded by $|S_\Omega|$. Since $f^*$ minimizes the $\loss_{KR}$, one can show that this bound is in fact tight, hence the equality. 

\begin{figure}
    \centering
    \begin{tabular}{ccc}
        \includegraphics[height=0.11\textheight]{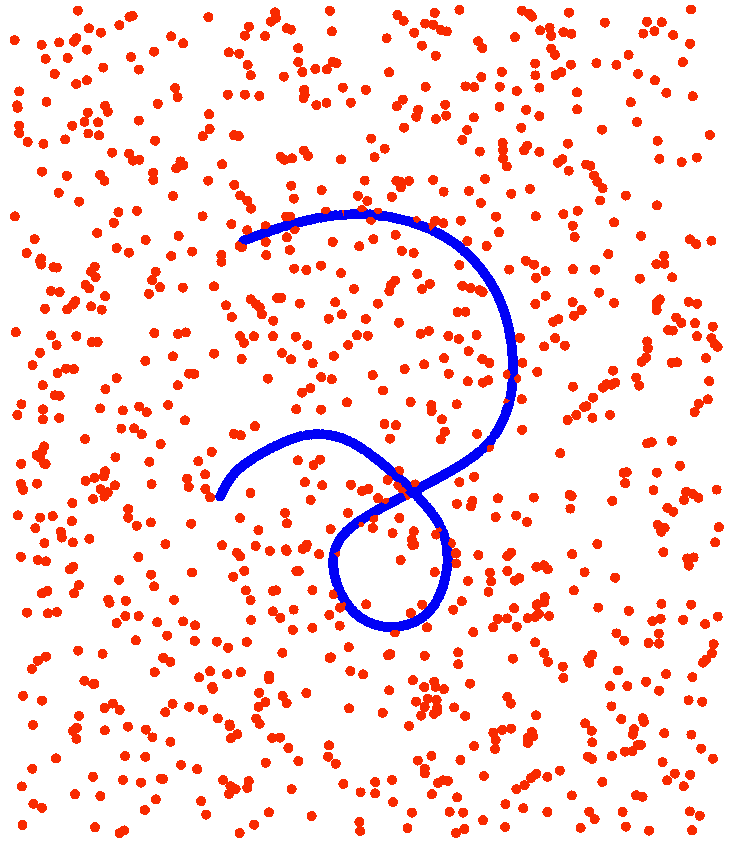} &
        \includegraphics[height=0.11\textheight]{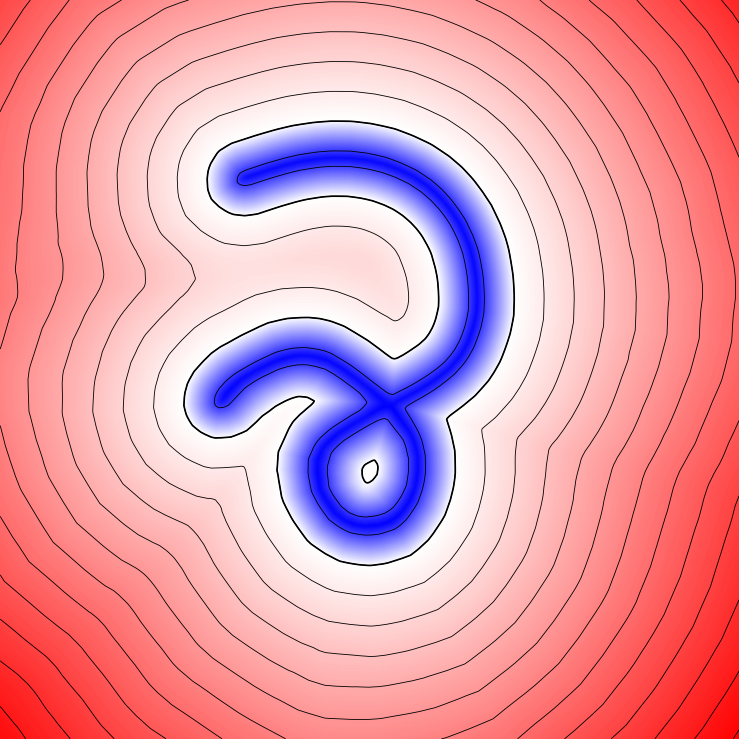} &
        \includegraphics[height=0.11\textheight]{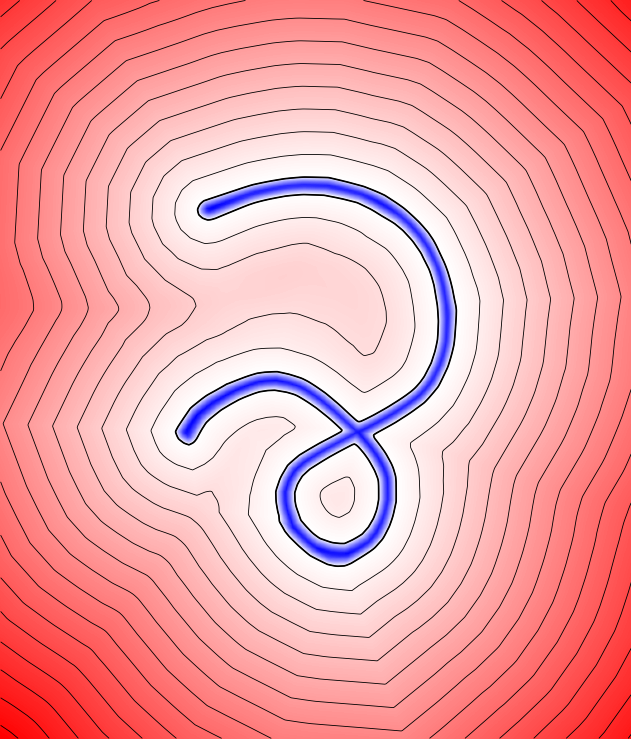} \\ 
        Input point cloud &
        $m=10^{-1}$ &
        $m=2 \times 10^{-2}$ \\
        
        \includegraphics[height=0.11\textheight]{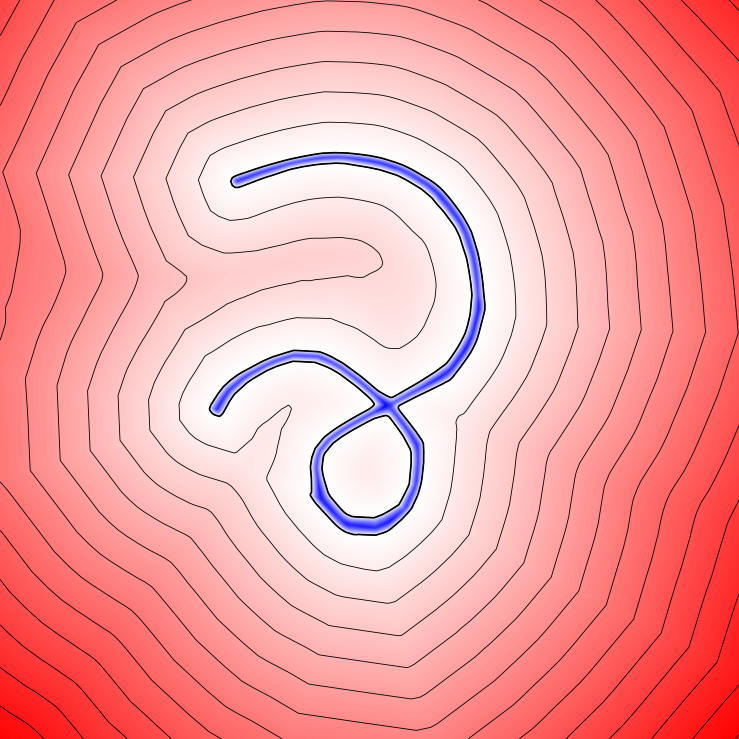} &
        \includegraphics[height=0.11\textheight]{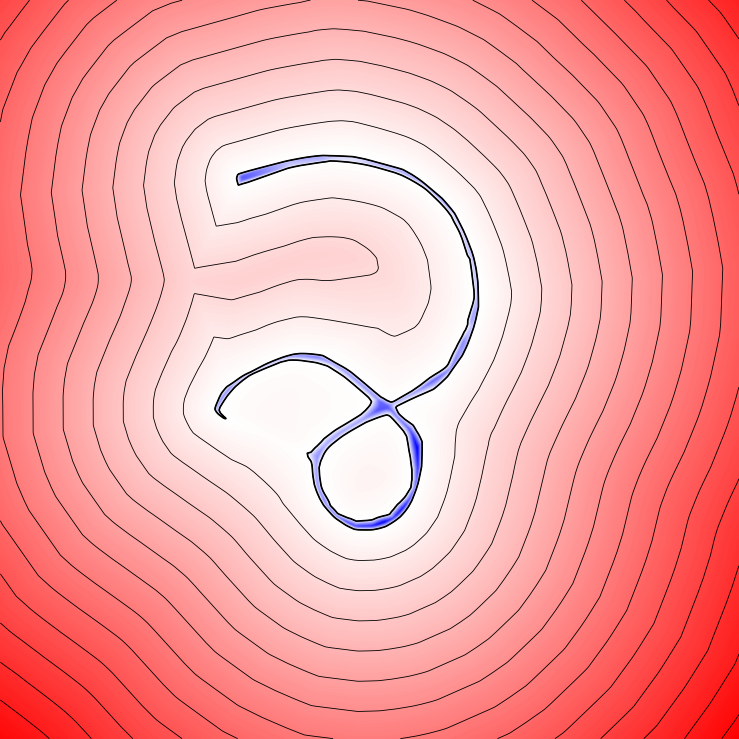} &
        \includegraphics[height=0.11\textheight]{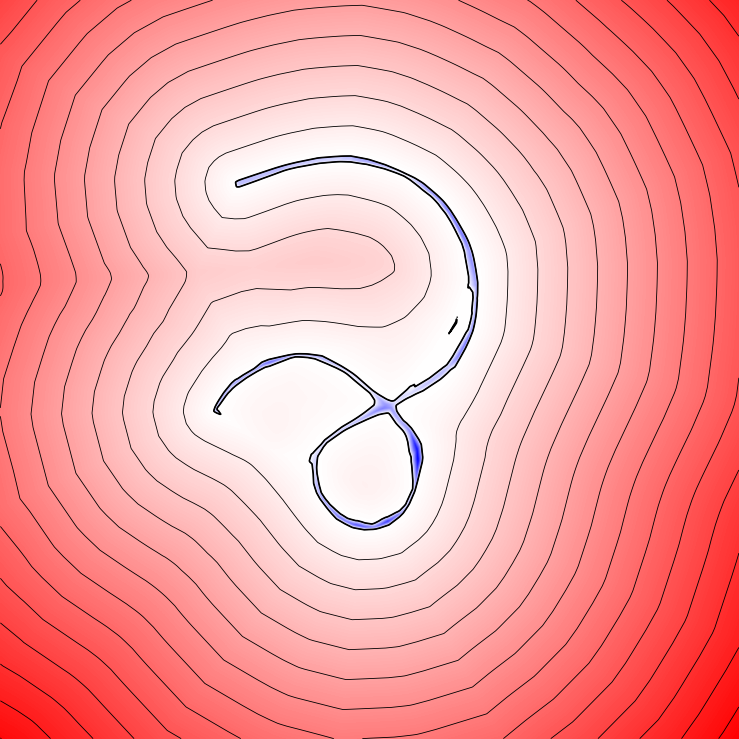} \\
        $m = 10^{-2}$ & 
        $m = 10^{-3}$ &
        $m = 10^{-4}$ \\
    \end{tabular}
    \caption{Neural distance field of an open curve in 2D. Minimizing $\loss_{hKR}$ on this dataset yields a distance field to the curve up to the margin parameter $m$. Large $m$ create large but consistent underestimation of the true distance while smaller $m$ lead to instabilities in training and final result. A value of $m$ around $10^{-2}$ is a good trade-off in practice.}
    \label{fig:influence_margin_unsigned}
\end{figure}

Minimizing $\loss_{hKR}$ over $1$-Lipschitz functions therefore approximates the signed distance function of the considered object. The critical quantity to control here is the margin parameter $m$, which can be thought as the minimal quantity to be imposed between points of different labels. As such, smaller margins allow the minimizer to better approximate the SDF $S_\Omega$. However, in the context of $1$-Lipschitz neural networks, the \emph{fat-shattering} dimension of the resulting function class has been shown to increase as $(\frac{1}{m})^n$~\cite{bethunePayAttentionYour2022}, which means that optimizing for smaller margins requires significantly more parameters in the network and leads to less stable results in practice (see for instance Figure~\ref{fig:influence_margin_unsigned}).  

The constraint on $\rho$, that is the exclusion of a shell of width $2m$ centered around the interface, means that the training dataset should ideally not contain any sample at distance smaller than $m$ from $\partial \Omega$. In practice, we do not explicitly prevent this from happening. As noticed by Béthune et al.~\cite{bethuneRobustOneClassClassification2023}, this introduces an additional error of the order of $m$. Finally, as far as the other parameter $\lambda$ is concerned, the hinge loss being a hard constraint in the theorem implies that $\lambda$ should be large enough in practical optimization.

Note that the minimization of the hKR loss is \emph{semi-supervised}: it is only necessary to know if a given point has been sampled from $D \backslash \Omega$ or from $D$ to learn a SDF and the ground truth $S_\Omega$ does not need to be known. This effectively turns the classical regression task of fitting a neural distance field into a classification task between inside and outside points. With this strategy, all we are left to do is to generate some dataset of points with associated binary $y$ labels to train a Lipschitz network, which boils down to determining if a given point $x \in \real^n$ belongs to $\Omega$ or not.

\begin{figure}
    \centering
    \includegraphics[width=\linewidth]{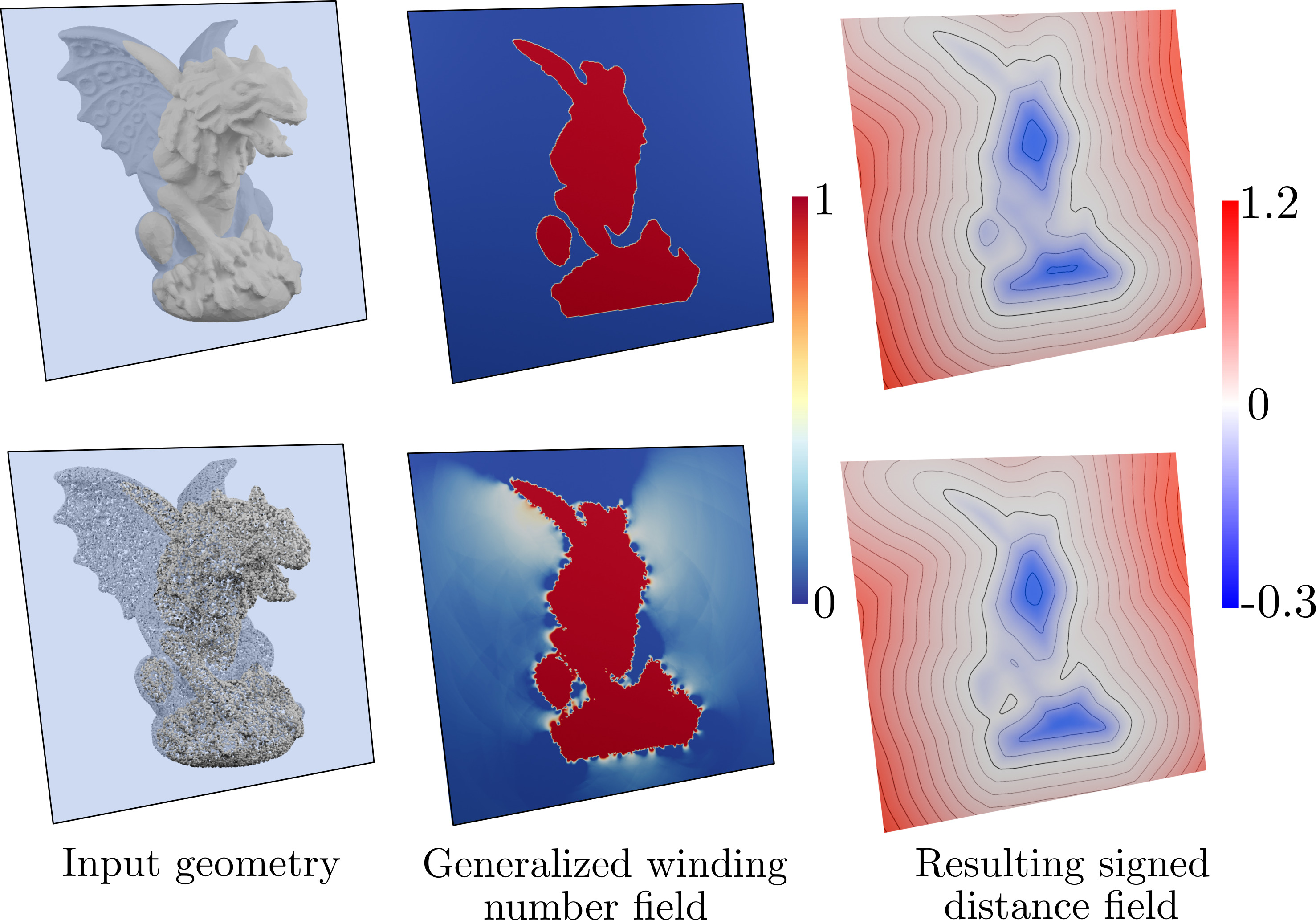}
    \caption{Generalized winding number field computed for the \emph{Gargoyle} model on the original manifold mesh (top) and a point cloud of 50K points(bottom). Thresholding this field allows to partition a dataset of points into inside and outside of the shape, from which a neural signed distance field can be optimized by minimizing the hKR loss (right column).}
    \label{fig:gargoyle_winding_number}
\end{figure}

\begin{figure*}[h]
    \centering
    \begin{tabular}{c c c c c c c}
         \includegraphics[width=0.14\textwidth]{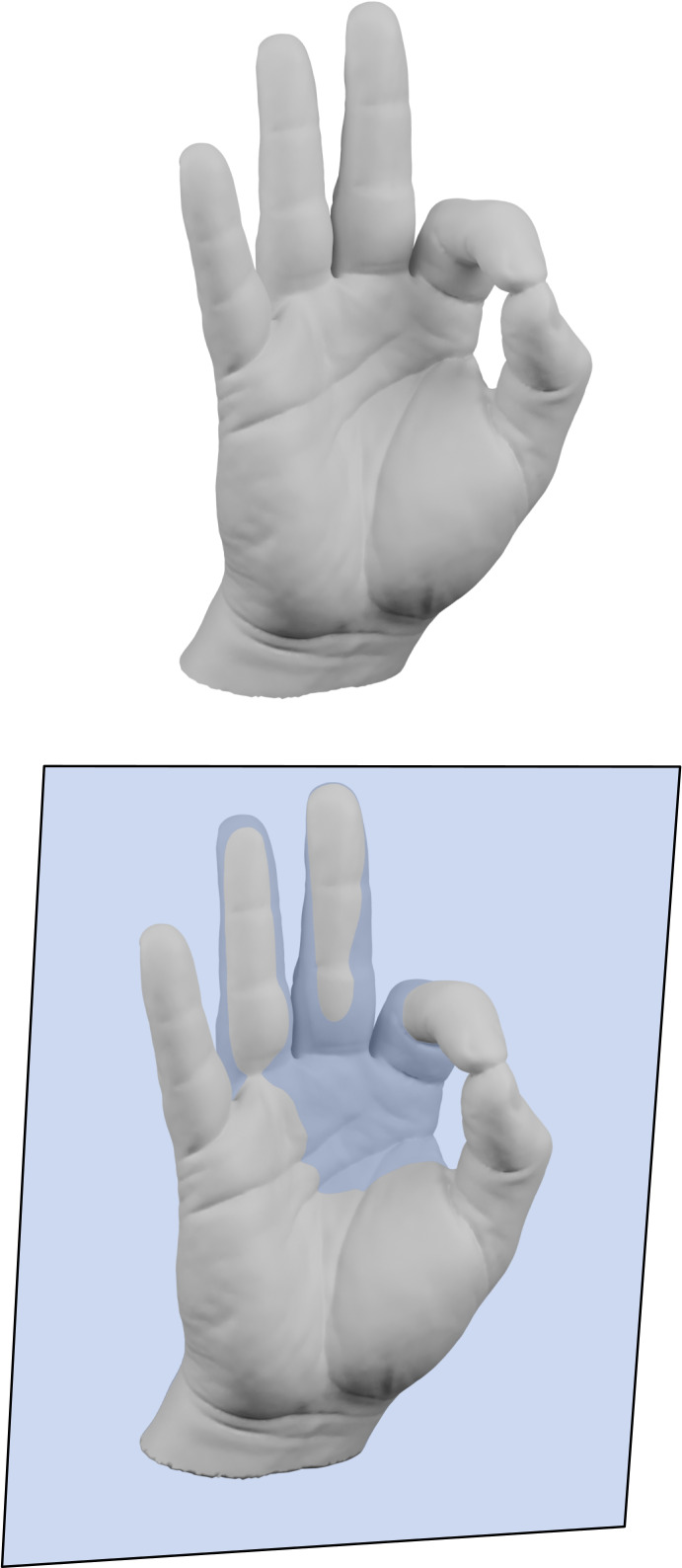} &
         \includegraphics[width=0.13\textwidth]{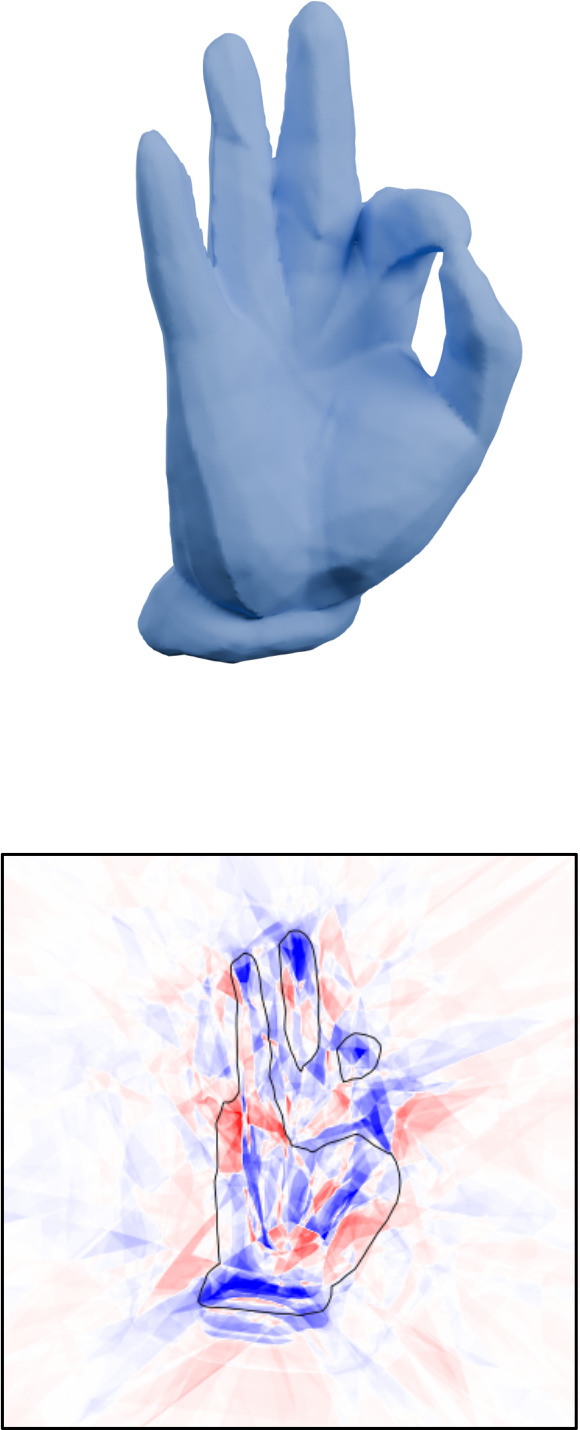} &
         \includegraphics[width=0.13\textwidth]{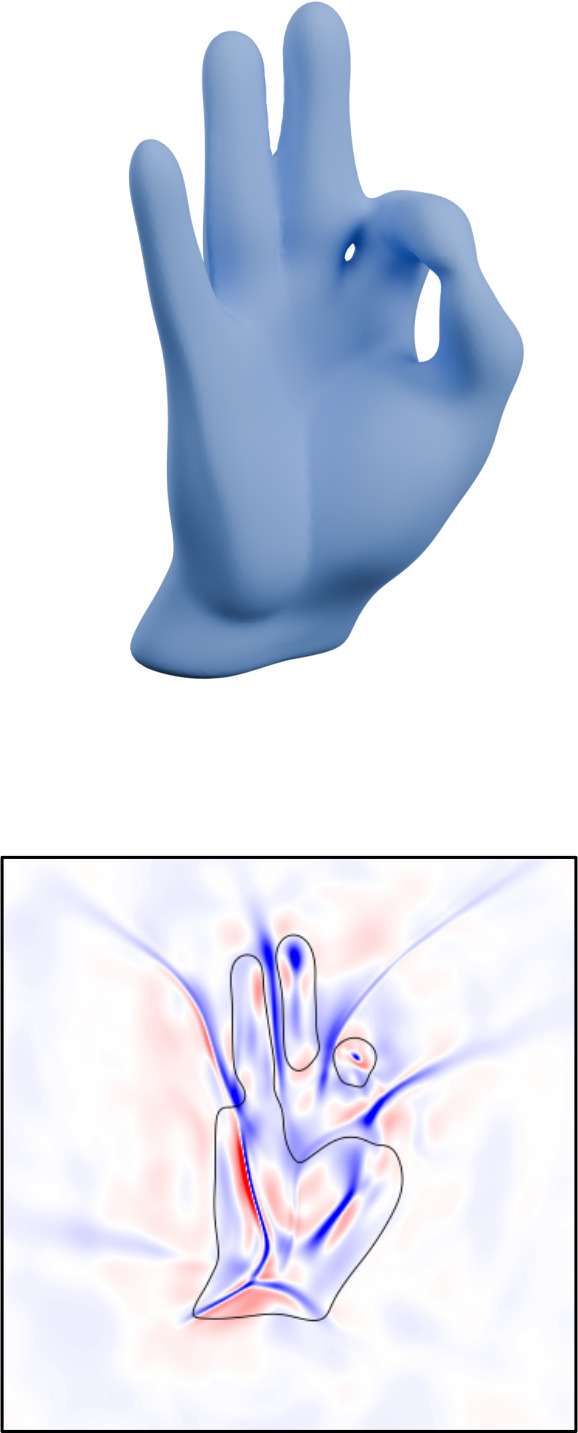} &
         \includegraphics[width=0.13\textwidth]{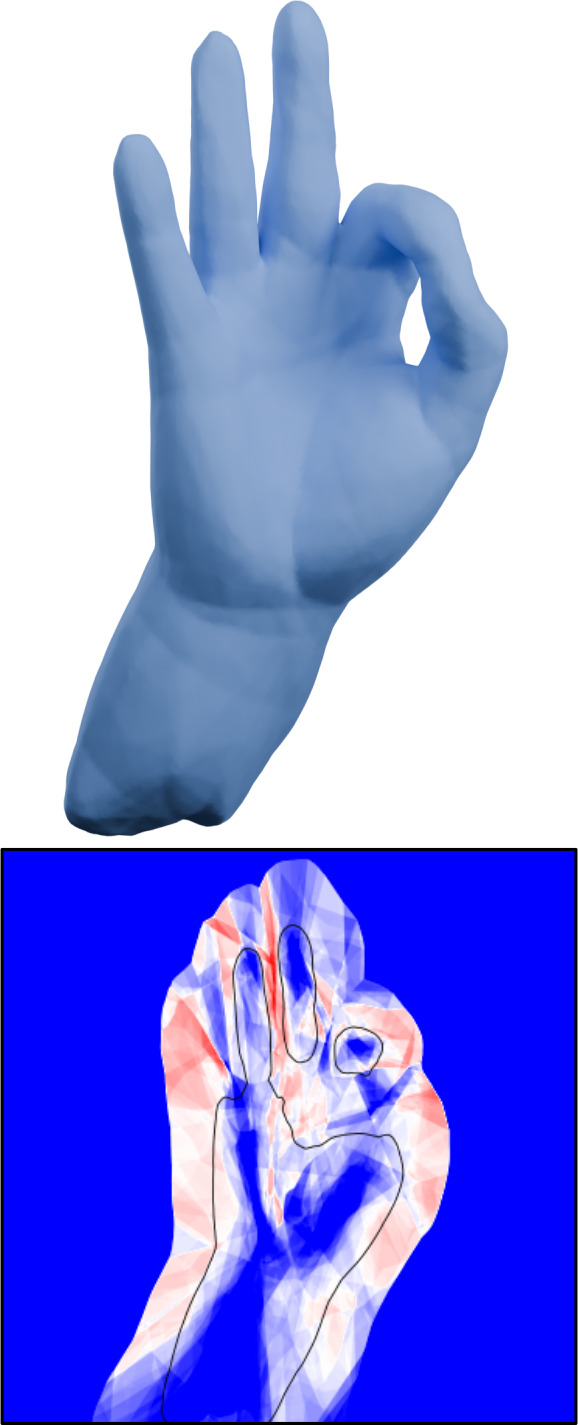} &
         \includegraphics[width=0.13\textwidth]{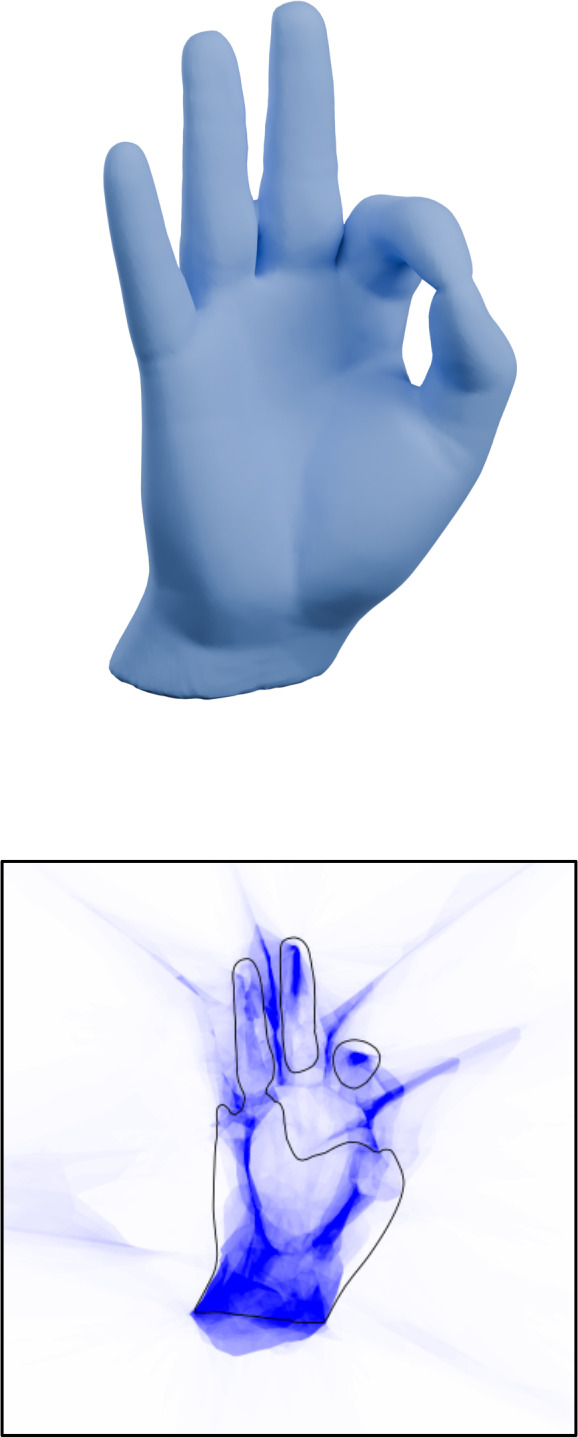} &
         \includegraphics[width=0.13\textwidth]{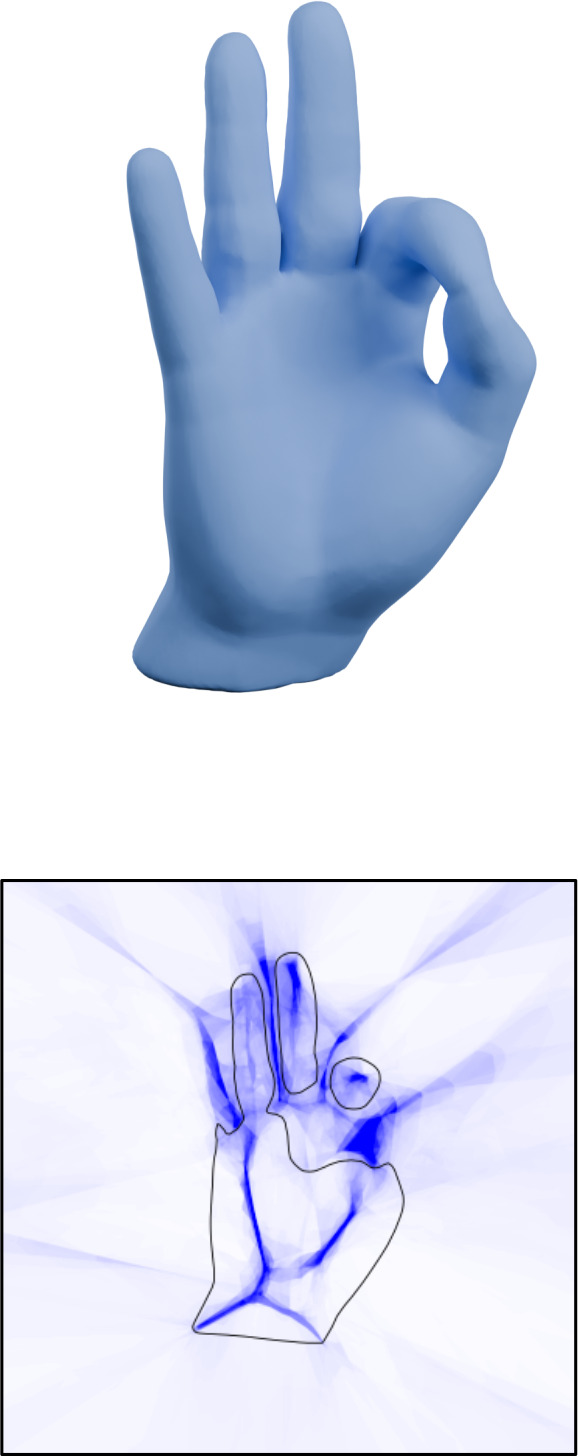} &
         \includegraphics[width=0.025\textwidth]{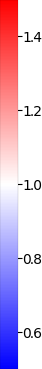} \\
         Ground Truth &
         ReLU MLP~\cite{daviesEffectivenessWeightEncodedNeural2021} &
         SIREN~\cite{sitzmannImplicitNeuralRepresentations2020} &
         SALD~\cite{atzmonSALDSignAgnostic2020a} &
         Ours, min $\loss_{fit}$ &
         Ours, min $\loss_{hKR}$ & \\
    \end{tabular}
    \caption{Comparison of various neural distance field methods on a hand dataset. Top row: reconstructed zero level set using the marching cube algorithm~\cite{lorensenMarchingCubesHigh1987}. Bottom row: norm of the gradient along a cut. While their reconstruction of the zero level set is comparable, previous methods all present a gradient norm that either exceeds one or vanishes far from the surface.}
    \label{fig:hand_reconstruction}
\end{figure*}

\subsection{Inside/Outside Partitionning}
\label{ssec:winding}

Let us first consider the case where the input $\partial \Omega$ represents a closed curve in the plane or a closed surface.  If the input geometry is clean enough, like for instance a manifold surface mesh, determining its inside from its outside is a task that can be solved in a robust way~\cite{sutherlandCharacterizationTenHiddenSurface1974}. When the geometry presents holes, defects, or is simply made of points, a notion of "insideness" can still be recovered by computing the \emph{generalized winding number}~\cite{jacobsonRobustInsideoutsideSegmentation2013}. Intuitively, the winding number $w_\Omega$ of a surface $\partial \Omega$ at point $x$ is the sum of signed solid angles between $x$ and surface patches on $\partial \Omega$. For a closed smooth manifold, the values amounts at how many times the surface "winds around" $x$, yielding an integer value. When computed on imperfect geometries, $w_\Omega$ becomes a continuous function (see Figure~\ref{fig:gargoyle_winding_number}). Through careful thresholding, it is still possible to determine points that are inside or outside the shape with high confidence.

Going back to our problem of learning a signed distance function from surface data $\partial \Omega$, we first sample points $X$ uniformly inside a loose bounding box $D$ of $\Omega$, and compute their generalized winding number using the method of Barill et al.~\cite{barillFastWindingNumbers2018}. Then, two distributions $X_{in}$ and $X_{out}$ are extracted from $X$ by choosing $N$ points from $X$ with winding number smaller than $\tau_i$ and greater than $\tau_o$ respectively, where $\tau_o < \tau_i$ are threshold parameters. This ensures that the network will be trained using the same amount of interior and exterior points. In practice, $N$ varies from $10^4$ to $10^6$ points, depending on the amount of details needed to be captured. Points from $X_{out}$ are assigned a label $y=1$ and points from $X_{in}$ a label $y=-1$, before being fed in a Lipschitz neural network that minimizes the hKR loss (Equation~\eqref{eq:hkr}). As shown on the right of Figure~\ref{fig:gargoyle_winding_number}, this process effectively allows to recover a signed distance function of the original shape.

\subsection{Distance field of shapes without interior}
\label{ssec:no_interior}

If the input object $\Omega$ has no interior, like the case of a curve or an open surface, we can still learn a distance function to $\Omega$ by minimizing the hKR loss without relying on the generalized winding number. We simply sample $X_{in}$ as being points on the surface or curve, either by taking directly the point cloud as input or sampling on triangles. $X_{out}$ is then a uniform distribution over the domain $D$. Up to the margin parameter $m$ in the loss, this still results in an approximation of the distance function of the considered manifold. But since level sets of the neural function can only be closed surfaces, the optimization results in an object having a "thickness". This property is directly controlled by the margin parameter $m$ in the loss, as shown in Figure~\ref{fig:influence_margin_unsigned}: a larger margin leads to reconstruction of the data with a non-negligible thickness, whereas a margin too small leads to instabilities on the final result. Figure~\ref{fig:bettle_knot} further demonstrates the SDF reconstruction ability of our method in three dimensions in the context of open surfaces or curves. Note that this setup is also suitable for closed surfaces in order to learn an unsigned distance field, as it is the case in Figure~\ref{fig:overview}.
\section{Results and Applications}

\subsection{Implementation details}

We implement our method in python using the \emph{Pytorch} library for neural network training. All our experiments were performed on a Ubuntu 22 workstation using a Nvidia 4070Ti GPU. For the generalized winding number, we use the original implementation of Barill et al.~\cite{barillFastWindingNumbers2018} as provided in \emph{libigl}~\cite{libigl}.

We use a neural network architecture of 20 SLL layers of size $k=128$. In total, this amounts for 330K floating point parameters for a total size of approximately 1.4MB. For reference, the hand mesh used in Figure~\ref{fig:hand_reconstruction} is described by 150K floating point numbers for vertices and 300K integers for faces, for a total size of 2.2MB when compressed.

All inputs are normalized in a bounding box $[-\frac{1}{2} ; \frac{1}{2}]^n$ before any processing. We set $D = [-1,1]^n$ as our sampling domain. With the exception of Figure~\ref{fig:influence_margin_unsigned}, all our experiments were performed with $m=10^{-2}$ and $\lambda=100$. 

\subsection{Gradient Robustness}

\begin{figure}
    \centering
    \begin{tabular}{cc}
        \includegraphics[height=0.26\textheight]{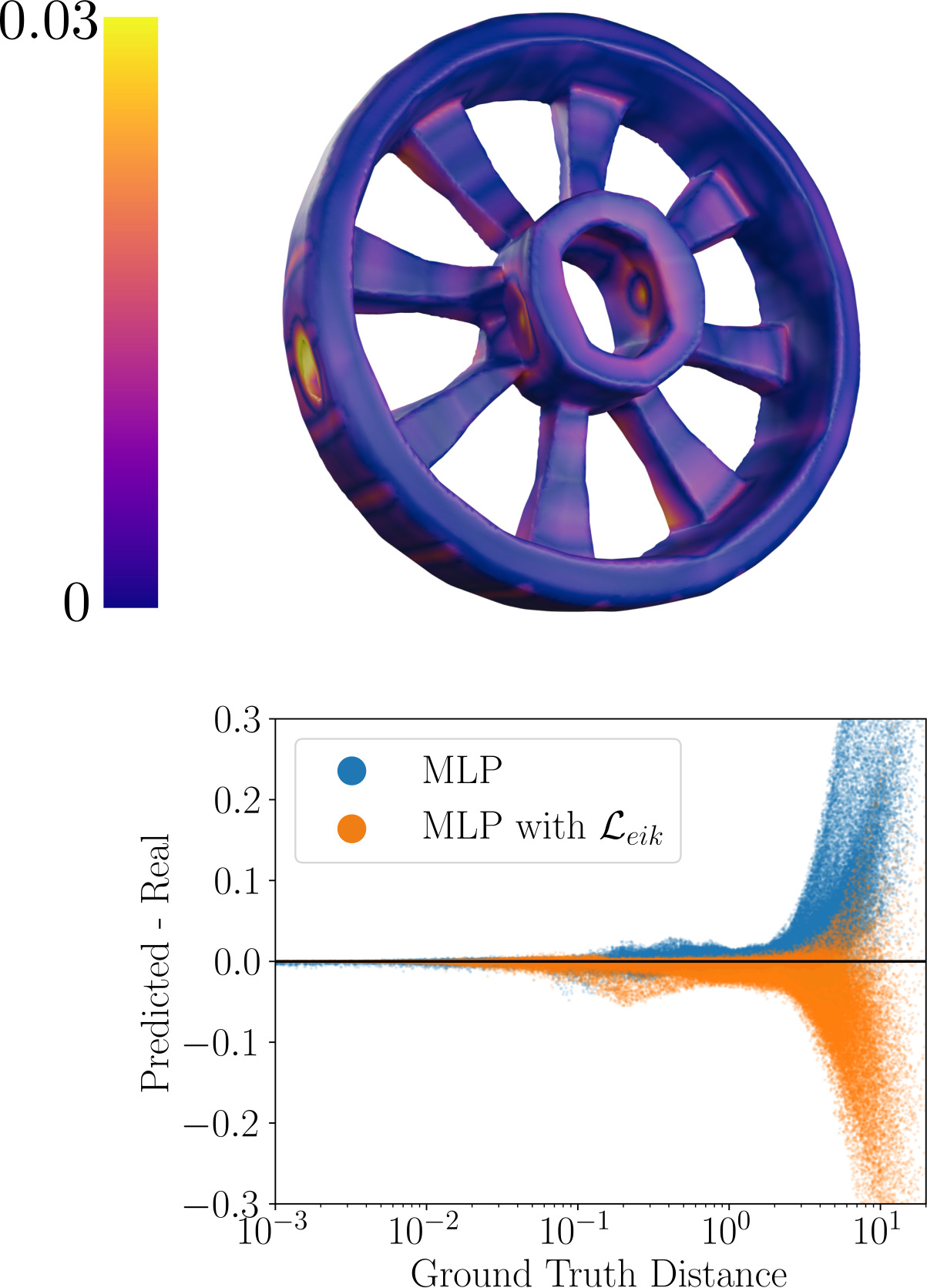} & 
        \includegraphics[height=0.26\textheight]{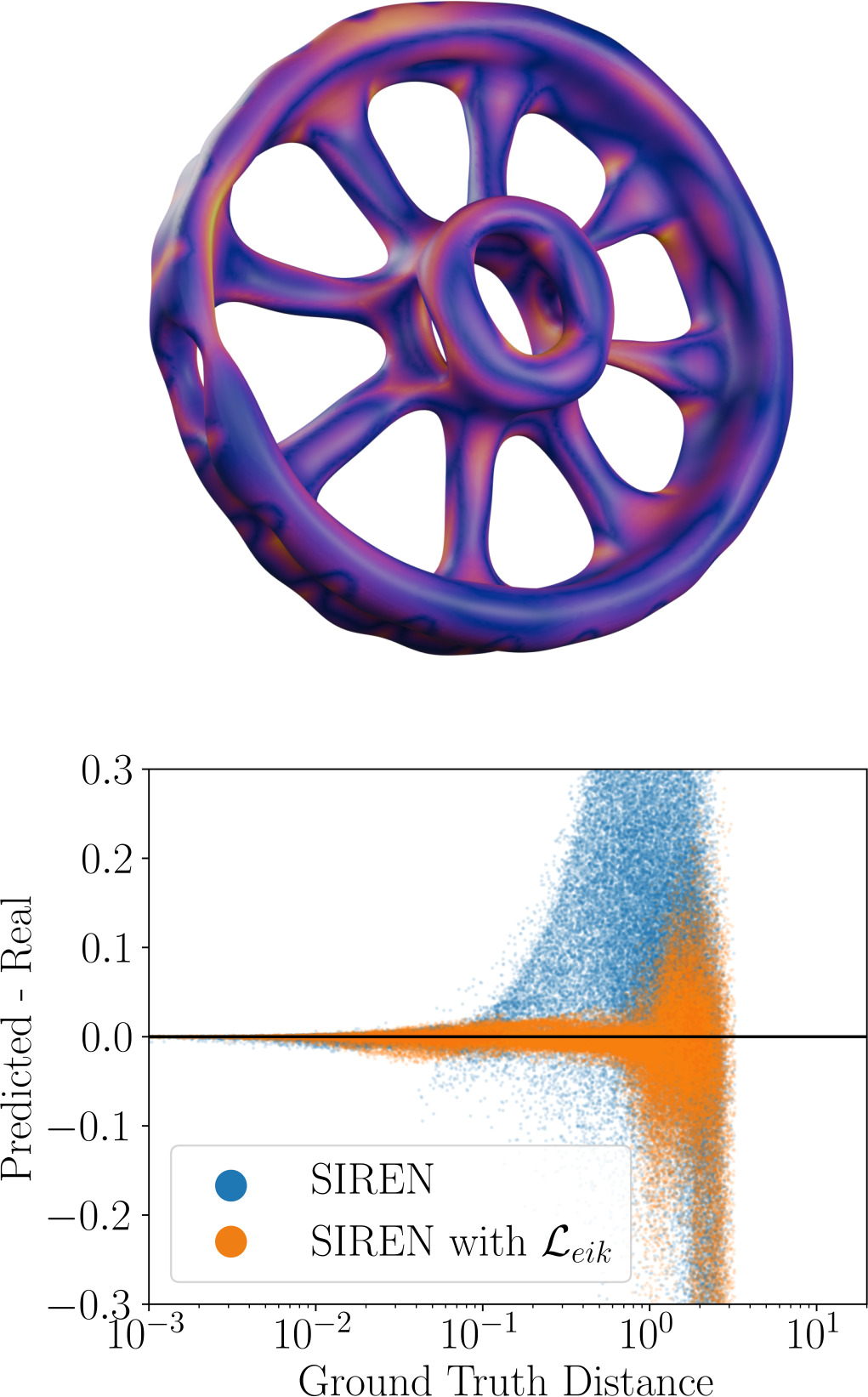} \\
        ReLU MLP~\cite{daviesEffectivenessWeightEncodedNeural2021} &
        SIREN~\cite{sitzmannImplicitNeuralRepresentations2020} \\
        \includegraphics[height=0.26\textheight]{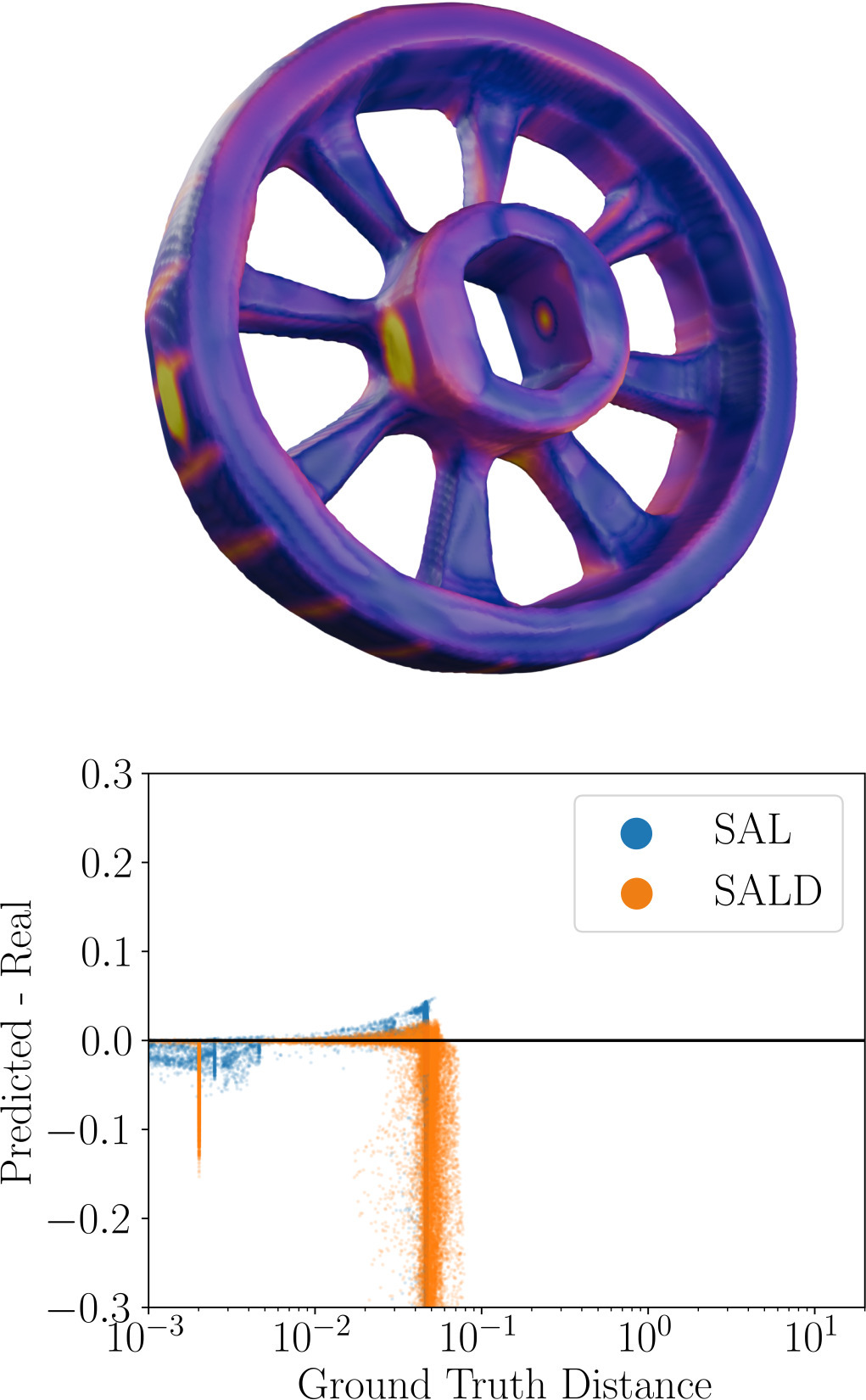} & 
        \includegraphics[height=0.26\textheight]{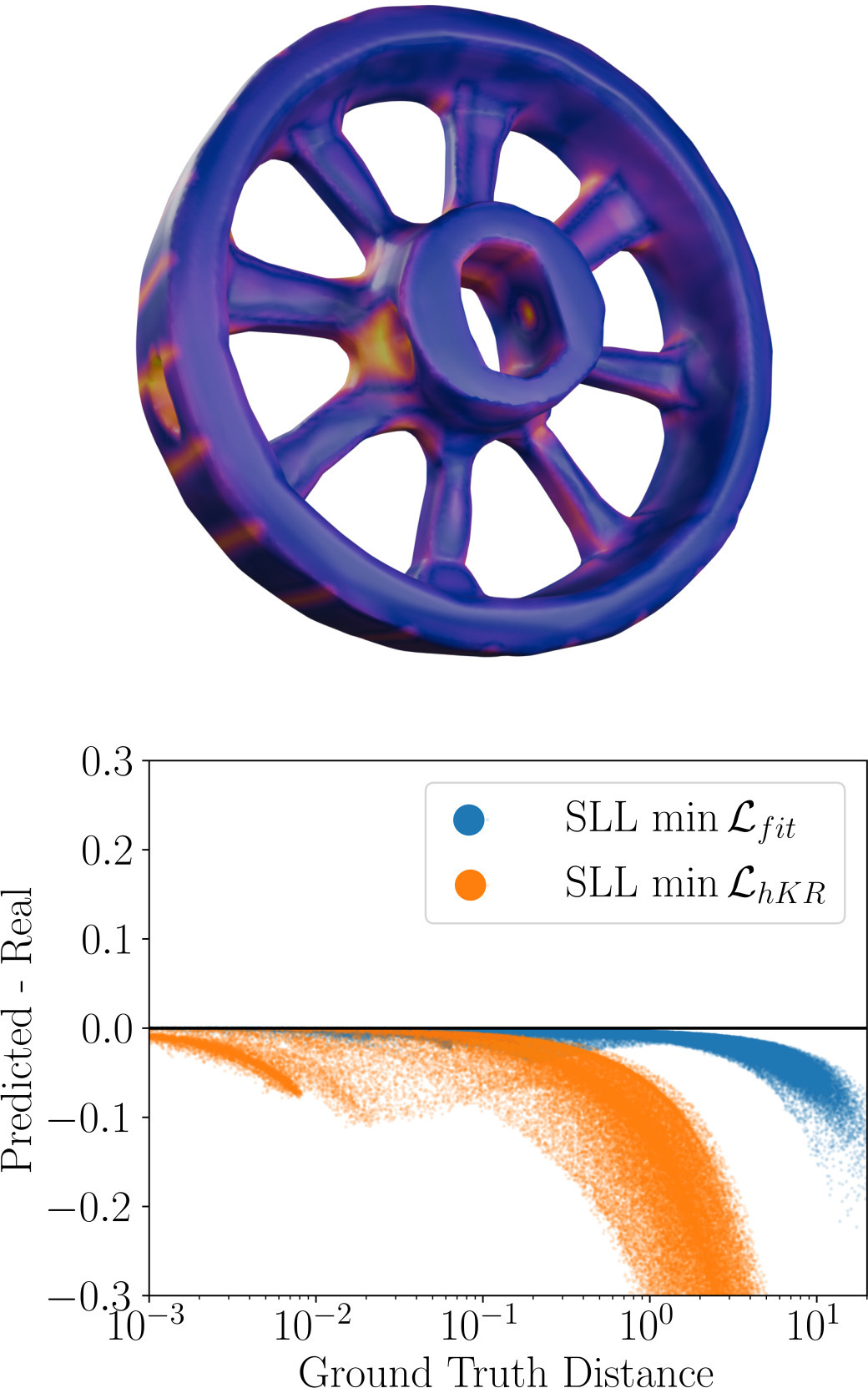} \\
        SAL~\cite{atzmonSALSignAgnostic2020a} - SALD~\cite{atzmonSALDSignAgnostic2020a} &
        Ours \\
    \end{tabular}
    \caption{Reconstruction experiment on a handle mesh. We plot the zero level set of each neural network with a colormap indicating the distance from the ground truth's surface. Our results around the zero level set are comparable to other methods. Moreover, we evaluate the difference $f_\theta(x) - S(x)$ for 100K points sampled at random and plot the results. As expected, this difference is always negative for our method, indicating that we always underestimate the true distance.}
    \label{fig:handle_distance}
\end{figure}

As a first experiment, we demonstrate the robustness of our approach in comparison with neural distance field methods from the state of the art. These methods are trained on a dataset of points $(x_i, S_\Omega(x_i))$ with $S_\Omega$ being precomputed from the input mesh. They minimize a least-square fitting loss:
\begin{equation}
    \loss_{fit} = \sum_i \left[f_\theta(x_i) - S_\Omega(x_i) \right]^2
\label{eq:loss_fit}    
\end{equation}
to match the true distance function, as well as an eikonal loss:
\begin{equation}
    \loss_{eikonal} = \int_D (||\nabla f_\theta(x)|| - 1)^2 \, dx.
    \label{eq:loss_eikonal}
\end{equation}
that regularizes their Lipschitz constant. For a fair comparison, the architecture reported by the original works are scaled to match our number of parameters.

On Figure~\ref{fig:hand_reconstruction}, we reconstruct the zero level set of the neural function using marching cubes and plot its gradient norm over some planar cut. We observe that the classical multilayer perceptron (MLP) with ReLU activations and SIREN~\cite{sitzmannImplicitNeuralRepresentations2020}, which is a MLP with sine activations, are able to capture the surface and the topology of the input, yet their gradient is unstable and often exceeds unit-norm. SALD~\cite{atzmonSALDSignAgnostic2020a} is a method that fits a neural implicit surface without requiring knowledge of the sign of its distance function. However, the trained network has a null gradient far from the surface. As far as our method is concerned, we observe that when $\loss_{fit}$ (Equation~\eqref{eq:loss_fit}) is minimized over a Lipschitz architecture, the zero level set represents the input shape more accurately than with the hKR loss. This behavior is expected since the network has access to more information in the dataset, and also because of the slight error introduced by the margin $m$ (as discussed in Section~\ref{ssec:hkr}).

\subsection{Signed vs Unsigned Distance Field}
\label{ssec:signed_vs_unsigned}

\begin{figure}
    \centering
    \includegraphics[width=\linewidth]{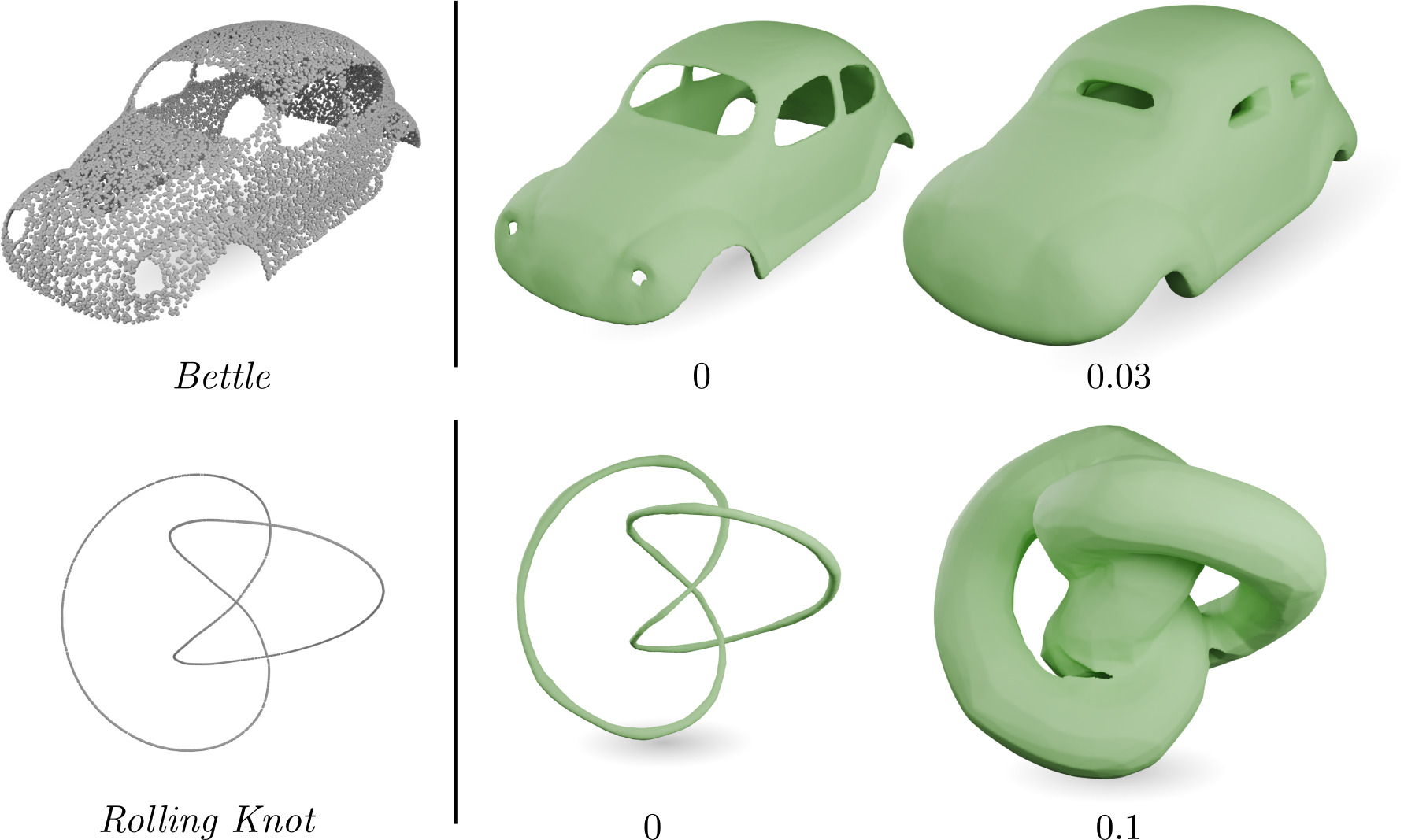}
    \caption{Learning of an unsigned distance field in the case of  an open surface with holes (top) and a curve in 3D (bottom) represented as point clouds. We display level-set extracted via the marching cube algorithm for two values of the distance function. The neural field correctly captures the geometry of the object.}
    \label{fig:bettle_knot}
\end{figure}

As our method provides stable distance estimation far from the zero level set, we are also able to extract high quality isosurfaces for different values of the function. This is illustrated on Figure~\ref{fig:teaser} for the signed case (using the generalized winding number) and on Figure~\ref{fig:bettle_knot} in the unsigned case, for a curve and an open surface with holes. Additional results are available in supplemental material. 

While fitting an unsigned distance also works perfectly fine for a closed object, partitioning the points first with generalized winding number enables our method to benefit from its robustness to faulty and noisy input. As a result, some defects of the zero level set can be repaired, as it is the case with the corrupted \emph{botijo} model shown in Figure~\ref{fig:overview}.

\subsection{Underestimation of the true distance field}

To further justify our claim that our learned signed distance field never overestimates the true distance, we train different neural distance fields on a \emph{handle} model (Figure~\ref{fig:handle_distance}). We then extract the zero level set of each method using marching cubes~\cite{lorensenMarchingCubesHigh1987} and sample 100K points uniformly in a sphere of radius 30 centered at zero. We plot the difference between the true distance $S(x)$ to the zero level set mesh and the predicted value of the neural network, in function of $S(x)$. In this experiment, a perfect network would output a straight line, but all methods present some deviation in practice. However, we observe that only our method provide only negative values, meaning that the network's output is always smaller than the true distance.

\subsection{Geometrical Queries}
\label{ssec:result_queries}

\begin{figure}
    \centering
    \includegraphics[width=\linewidth]{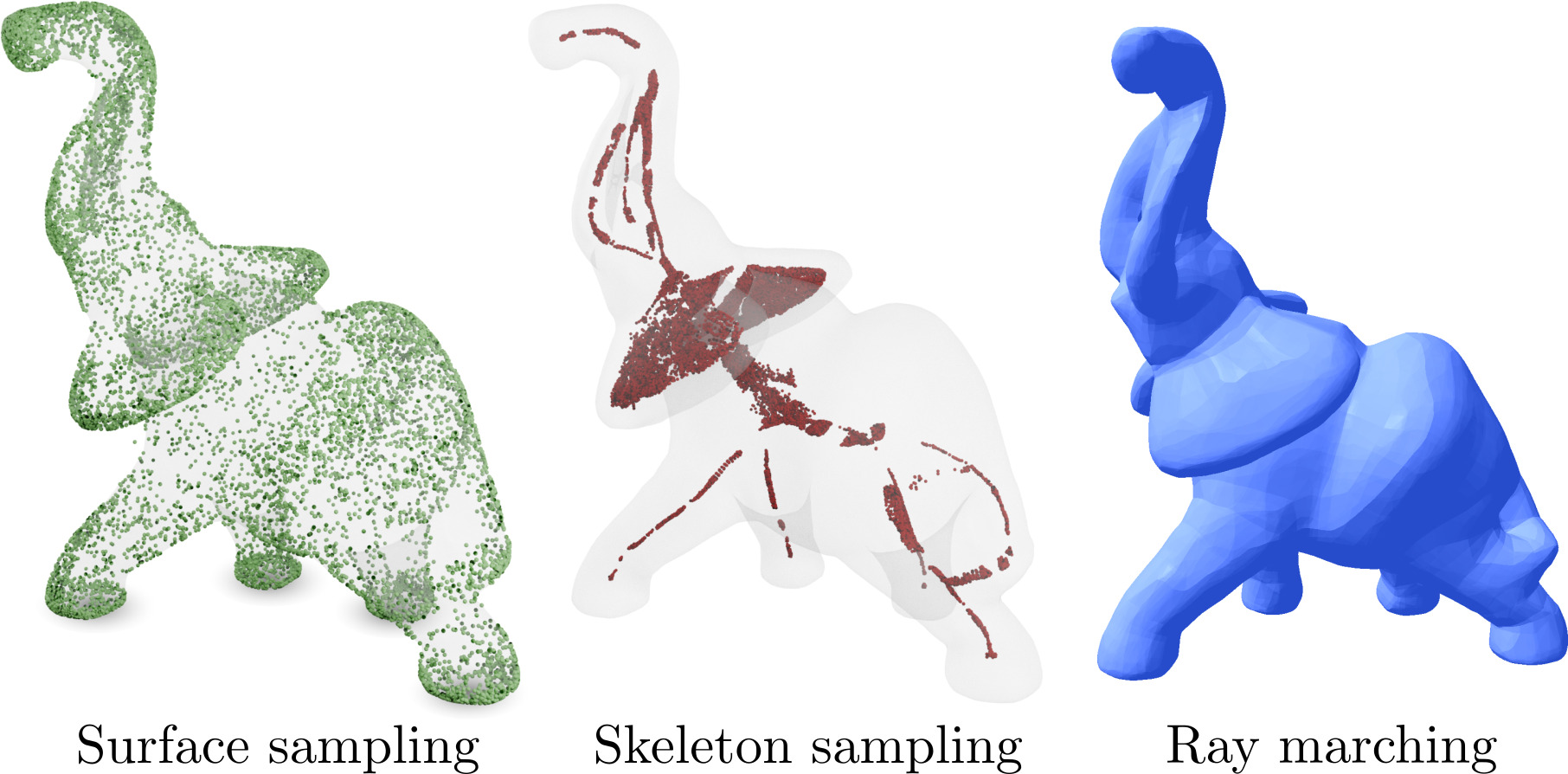}
    \caption{Geometrical queries performed on the \emph{elephant} model. Queries are all valid without relying on range analysis.}
    \label{fig:elephant}
\end{figure}

Having a robust SDF far from the surface enables efficient and robust geometrical queries on any of its level sets. In this section, we demonstrate experimentally that it is indeed possible to do on our Lipschitz network trained with the hKR loss, without requiring neither range analysis nor an estimation of the Lipschitz constant.

Given an exact SDF $S$ and a point $x$ in space, the closest point of $x$ on the zero level set of $S$ can be directly computed as $x - \nabla S(x) S(x)$. This expression falls short of the level set in the case where $||\nabla f|| < 1$ but can be iteratively evaluated until a convergence criterion is met. We illustrate this process on Figure~\ref{fig:elephant} (left), where points on a sphere of radius 2 are projected onto the surface of the \emph{elephant} dataset. As points tend to accumulate in salient regions of the model, a more uniform sampling can be recovered with the Iso-points method~\cite{yifanIsoPointsOptimizingNeural2021}.

Another application of neural distance fields is to sample the medial axis of an object. As the medial axis is defined as the set of points where the closest point on the surface is not unique, it corresponds to discontinuities of the gradient of the SDF. In our case, the Lipschitz network is fully differentiable and approximates the discontinuity by dropping the gradient's norm, as illustrated in Figure~\ref{fig:eikonal_cest_dla_merde} (right). This gives us a sample-and-reject strategy to construct the model's skeleton by evaluating the norm of the gradient. A mesh of the medial axis can then be extracted by the method of Clemot and Digne~\cite{clemotNeuralSkeletonImplicit2023}, whose skeleton sampling results we reproduce (Figure~\ref{fig:elephant}, middle) without any gradient regularization loss.

Finally, \emph{ray marching} queries the implicit function along a ray to determine the maximum step size that guarantees non-intersection. Without our $1$-Lipschitz network, we are able to take a step size of $1$ and recover a correct image (Figure~\ref{fig:elephant}, right).

\subsection{Constructive Solid Geometry}
\label{ssec:CSG}

\begin{figure}
    \centering
    \includegraphics[width=\linewidth]{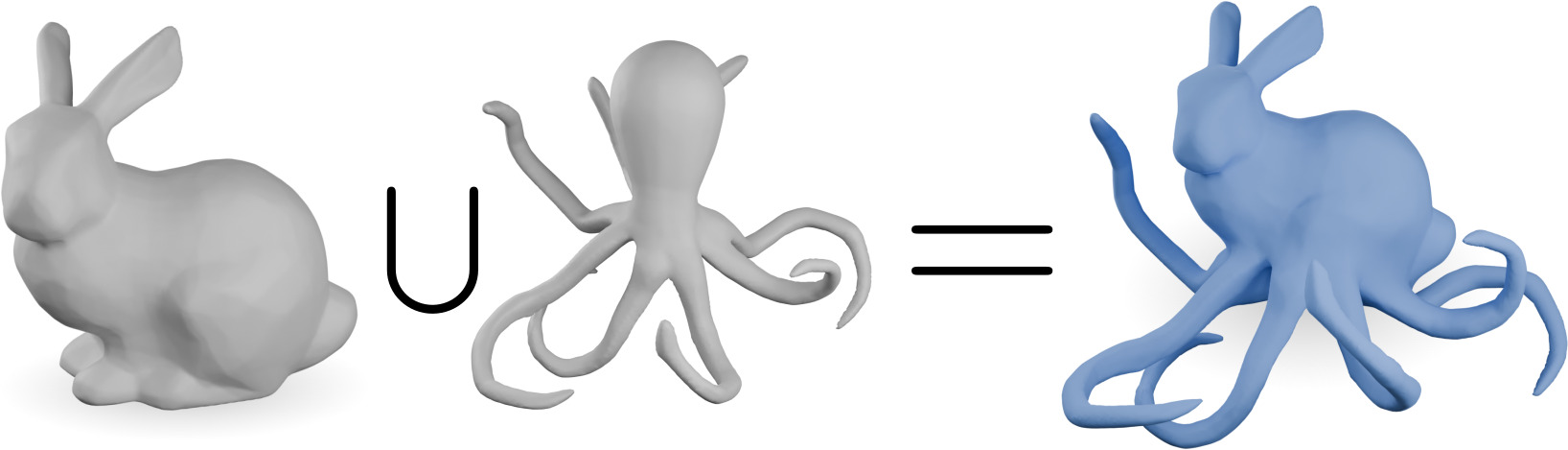}
    \caption{Constructive Solid Geometry. The union of two SDFs can be computed using the $\min$ operator. Although not necessarily a true SDF, the resulting function remains $1$-Lipschitz.}
    \label{fig:csg_octobunny}
\end{figure}

A key feature of implicit geometries is the ease with which we can apply boolean operations on their function to represent intersections, unions or differences of shapes~\cite{ricciConstructiveGeometryComputer1973}, thus building more complex shapes for simple ones. We illustrate this property on our neural distance field on Figure~\ref{fig:csg_octobunny}: given two SDFs $f_1$ and $f_2$, we can compute the surface of their union as the 0-level set of $\min(f_1, f_2)$. Similarly, their intersection coincides with the 0-level set of $\max(f_1,f_2)$. These operations preserve the Lipschitz constant so the guarantee of being $1$-Lipschitz still holds for the composition. However these point-wise minimum and maximum are not necessarily SDFs themselves~\cite{QuilezInterior, marschnerConstructiveSolidGeometry2023}.  

\subsection{Robustness to noise and sparsity of input}
\begin{figure}
    \centering
    \includegraphics[width=0.9\linewidth]{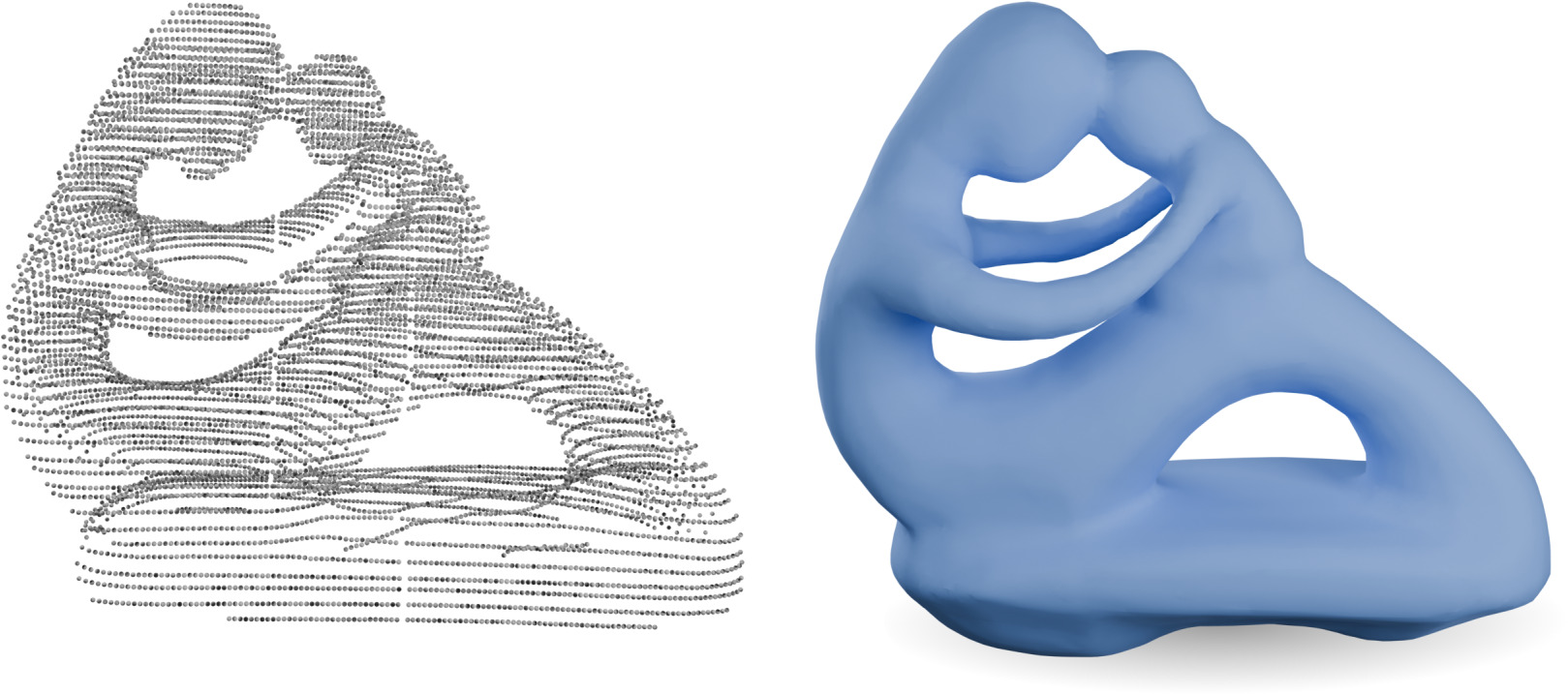} \\ 
    Simulated lidar~\cite{gschwandtner2011blensor} \\

    \includegraphics[width=0.9\linewidth]{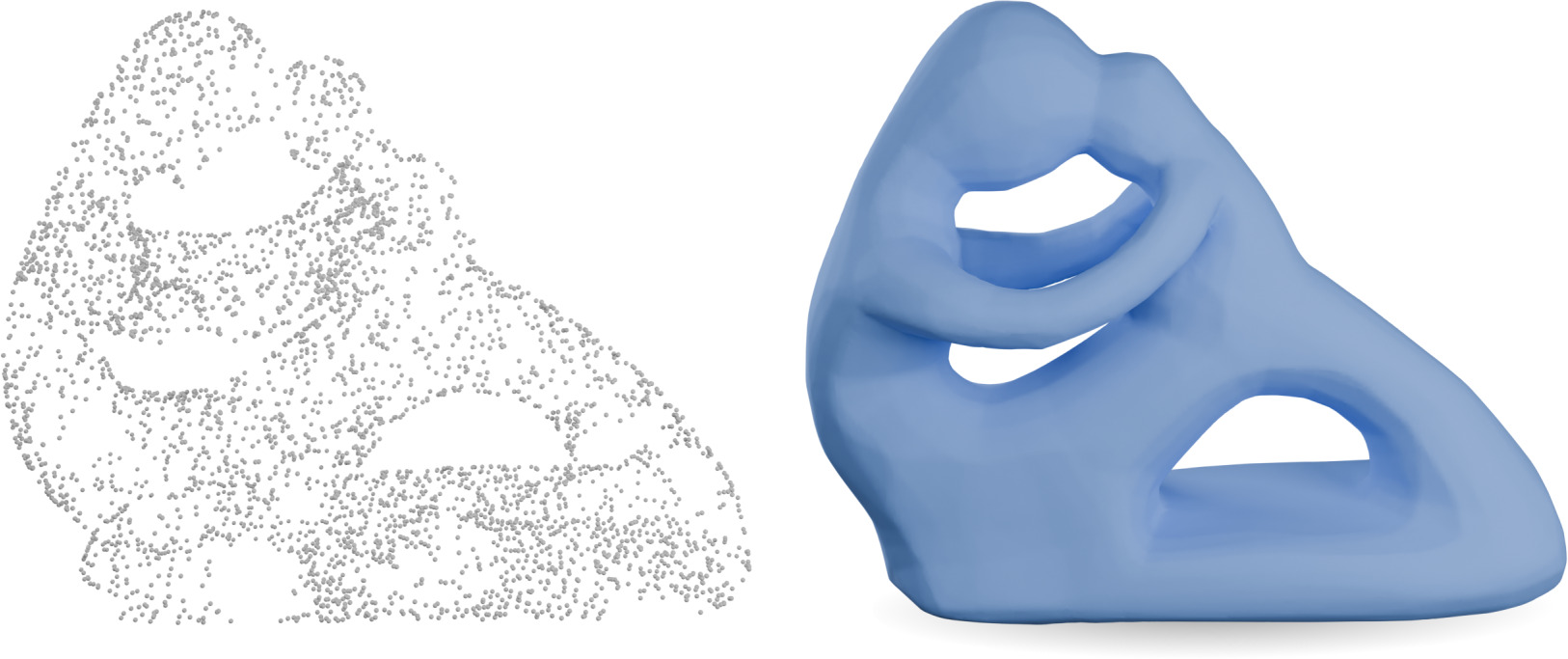} \\
    5000 points with holes \\

    \includegraphics[width=0.9\linewidth]{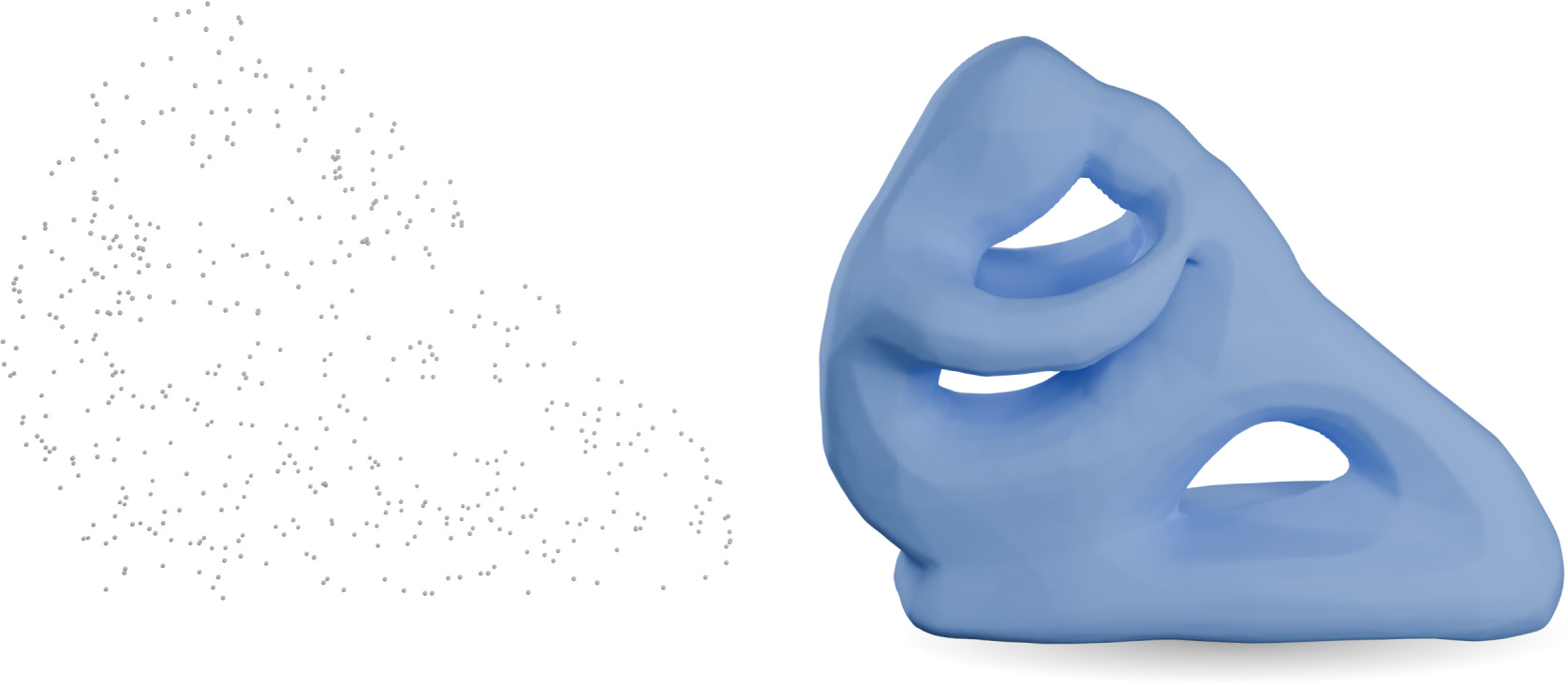} \\
    500 points \\

    \includegraphics[width=0.89\linewidth]{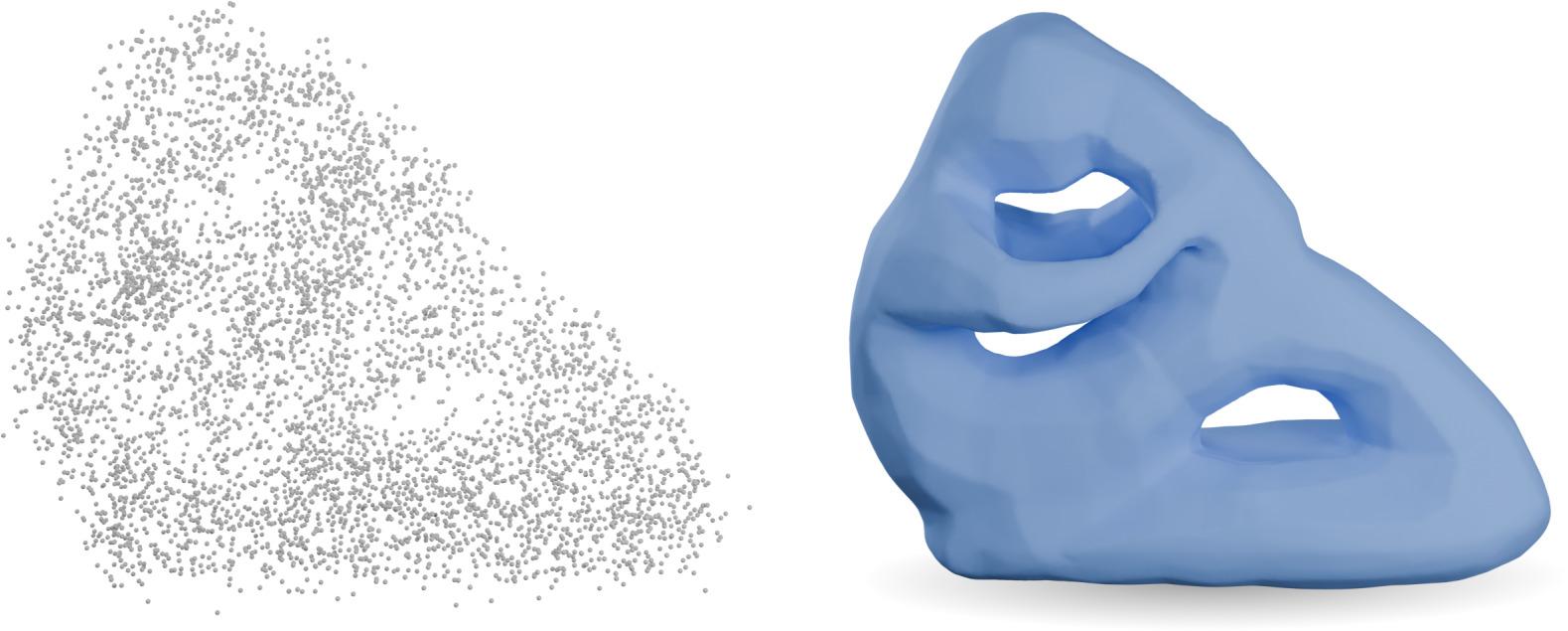} \\
    5000 points with gaussian noise
    
    \caption{Neural SDF reconstruction of the \emph{fertility} model in contexts where the ground truth is not available. Our method is able to recover the correct topology of the input surface even in very hard settings.}
    \label{fig:fertility}
\end{figure}

Finally, approaching the neural distance field problem with the hKR loss means that we are still able to recover a good approximation of the SDF of an object in contexts where the ground truth distance cannot be computed or is not available. We illustrate this on the \emph{fertility} dataset in Figure~\ref{fig:fertility} on four different settings. The first one is generated with \emph{Blensor}~\cite{gschwandtner2011blensor} and emulates a noisy Lidar acquisition of 7K points. The second one consists of 5K points where 50 points and their closest 40 neighbors have been removed from the point cloud, thus creating holes in the shape. The third one pushes sparsity to its maximum with only 500 points in total. Finally, the fourth one adds a Gaussian noise of standard variation $0.03$. In all four settings, the winding number still allows us to sample 10K inside and outside points. Minimizing the hKR loss then yields the correct topology on the zero level set, albeit some expected geometrical error. This also means that our method can be used as a noise reduction or an outlier elimination technique, a context where the hKR loss has already been used with success~\cite{bethuneRobustOneClassClassification2023}.

\section{Discussion and conclusion}

Neural distance fields are a promising technique for representing arbitrary geometry without relying on a discretization of space. While previous methods have demonstrated outstanding results in surface reconstruction and visual fidelity, we demonstrated that these neural functions could also be made $1$-Lipschitz and support geometrical queries without the need of range analysis. Having this $1$-Lipschitz constraint also allowed to learn a distance field without needing any distance information from the input shape, thus enabling neural fields applications for a broader set of geometry representations, even including bad quality point clouds or triangle soups.

Yet, this built-in robustness did not come without drawbacks. The hKR loss can only be successfully minimized up to a margin $m>0$, which prevents a perfectly accurate surface reconstruction. Smaller margin allows for more marginally more fine-grained details to be captured, but the resulting functions are usually biased towards low frequencies. Some recent neural methods solve this problem by using positional encoding at the start of the network~\cite{lipmanPhaseTransitionsDistance2021a}, like the \emph{Fourier features}~\cite{tancikFourierFeaturesLet2020}. A network using such an encoding could still be made $1$-Lipschitz: as the Lipschitz constant $K$ of the embedding can be known in closed form (it depends on the frequencies of the harmonic functions), it suffices to build a $1/K$-Lipschitz network by scaling the output. Yet, as the embedding does not preserve the gradient norm, so will the hKR-trained final network, which could lead to significant underestimation of the true distance. Improving reconstruction quality while being almost gradient preserving thus remains a challenge for future works.

Overall, for a fixed number of parameters, there seems to be a clear trade-off between quality of the zero level set and quality of the gradient far from it, an observation that has already been made in classification contexts regarding $1$-Lipschitz networks~\cite{bethunePayAttentionYour2022}. For a network of bounded depth, the zero level set is made of affine pieces whose maximum number grows as a polynomial of the width~\cite{telgarsky2016benefits,yarotsky2018optimal,bartlett2019nearly,piwek2023exact}, which would require a similar increase in numbers of parameters.

Finally, as Lipschitz networks are computationally heavier than their classical counterpart, future works could also focus on improving speed of convergence of the method, for example by reducing the total number of points needed in the dataset. This could be achieved by importance sampling before training, or by eliminating unnecessary points during training.

\appendix

\section{Proof of Theorem~\ref{thm:THE_THEOREM}}
\label{sec:proof_of_thm}

Let $f^*$ be a $1$-Lipschitz minimizer of $\loss_{KR}$, under the constraint that $\loss_{hinge}(f^*,y) = 0$. $f^*$ is shown to exist by Serrurier et al.~\cite[Theorem~1]{serrurierAchievingRobustnessClassification2021}. Define:

$$\partial \Omega^m = \{ x \in D ,\, |S_\Omega(x)| < m \}$$

as the "shell" of width $2m$ centered around $\partial \Omega$. Given the assumption on $\rho$, it exactly corresponds to the points that are not involved in the computation of the hinge loss. Then, the hinge loss being $0$ for $f^*$ implies that:

$$m - yf^*(x) < 0 \text{ for all } x \in D \backslash \partial\Omega^m.$$

In other words, for any point with distance larger than $m$ from the boundary $\partial \Omega$, $f^*$ has the same sign as $y$ and $|f^*(x)| \geqslant m$. Note that it would not be possible to have zero hinge loss if $\rho$ was not zero in $\partial \Omega^m$.

Let us first focus on what happens inside $\partial \Omega^m$. Since $f^*$ is $1$-Lipschitz, we have $|f^*(x)| \leqslant m$ for any $x \in \partial \Omega^m$.  Indeed, if it were not the case, let $x_0 \in \partial \Omega^m$ such that $f^*(x_0) > m$. Let:

$$z_0 \in \argmin_{S_\Omega(z) = -m} ||x_0-z||$$ be a projection of $x_0$ onto the $-m$ isosurface of $S_\Omega$. We have $f^*(z_0) \leqslant -m$ by continuity of $f^*$. By definition of $\partial \Omega^m$ and the Lipschitz property of $f^*$, we have:

$$|f^*(x_0) - f^*(z_0)| \leqslant ||x_0 - z_0|| \leqslant 2m$$

but also $f^*(x_0)>m$ and $f^*(z_0) \leqslant m$ which means that:

$$|f^*(x_0) - f^*(z_0)| = f^*(x_0) - f^*(z_0) > 2m$$

which is a contradiction. The case $f(x_0) < -m$ can be treated similarly by projecting onto the $+m$ isosurface of $S_\Omega$ and deriving the same inequalities.

In summary, the following inequalities hold:

$$\begin{array}{cc}
     |S_\Omega(x)| > m \implies& |f^*(x)| > m \\
     |S_\Omega(x)| \leqslant m \implies& |f^*(x)| \leqslant m  \\
\end{array}.$$

In particular, this means that inside $\partial \Omega^m$, the error made by $f^*$ is at most $2m$.

Now, consider the exterior of the shell. By the Lipschitz property of $f^*$, it is clear that $yf^*(x) \leqslant yS_\Omega(x)$ for all $x$. Suppose that this inequality is not tight for some $x_0 \in D \backslash \partial \Omega^m$. By continuity, this means that there exists a ball $B_\varepsilon$ included in $D \backslash \partial \Omega^m$ around $x_0$ for which the inequality is also strict. Computing the KR loss over the domain yields:

$$\int_D -\rho yf(x)dx  > \int_D -\rho y S_\Omega(x).$$

However, since $S_\Omega$ is a $1$-Lipschitz function and $\loss_{hinge}^m(S_\Omega, y) = 0$, this inequality is in contradiction with the fact that $f^*$ is a minimizer of $\loss_{KR}$. Combined with the fact that $f^*$ has the same sign as $y$ (and therefore as $S_\Omega$), we can conclude that:

$$\forall x, |S_\Omega(x)| > m \implies f^*(x) = S_\Omega(x).$$
\qedwhite



\bibliographystyle{eg-alpha-doi} 
\bibliography{biblio_zotero, biblio_other}

\end{document}


\vspace{-1cm}
\maketitle
\vspace{-1cm}

\section{Isosurface extraction}

In this section, we extract isosurfaces for different values of the distance on a larger variety of models. Classical models of geometry processing, namely the \emph{Stanbford bunny} and the \emph{bettle} model can be accessed here: \href{https://github.com/alecjacobson/common-3d-test-models?tab=readme-ov-file}{https://github.com/alecjacobson/common-3d-test-models?tab=readme-ov-file}. Models from the \emph{Rolling knot} onward are from the Thingi10K dataset~\cite{Thingi10K}. The \emph{Shirt} model is taken from the MGN dataset~\cite{bhatnagar2019mgn}.
The Gyroid is a noisy triangle soups of 2.9M faces with 3000 tiny connected components.

Signed distance fields (using the inside/outside partition of points) are depicted in blue, while unsigned distance fields are green. Interesting failure cases are the \emph{Wicker Chair} model, on which our method was not capable to capture all the details, and the \emph{Shirt} model, where holes at the end of sleeves are missing. We believe that both issues can be resolved by greatly increasing the number N of sampled points and the size of the Lipschitz network.

\includegraphics[width=\textwidth]{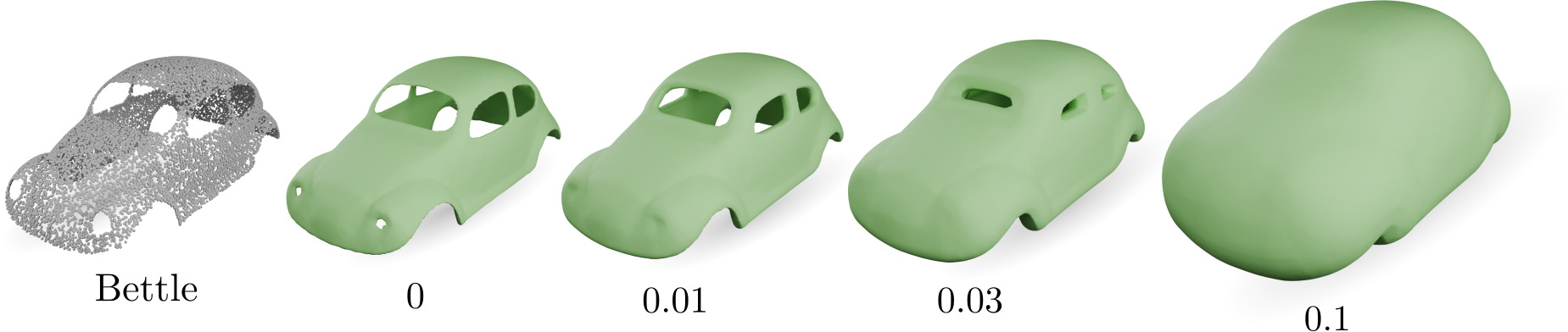} \\ ~ \\
\includegraphics[width=\textwidth]{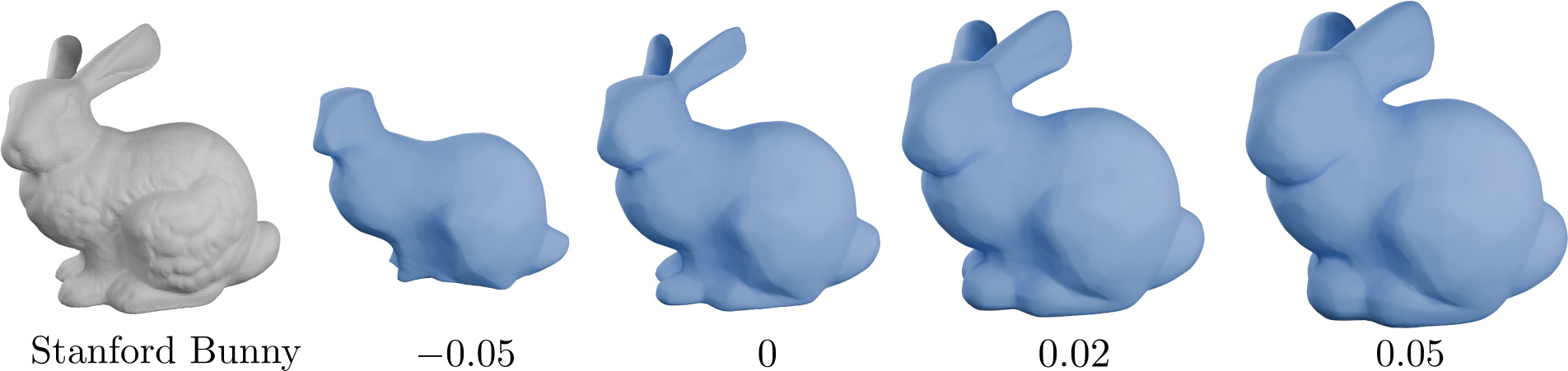} \\ ~ \\
\includegraphics[width=\textwidth]{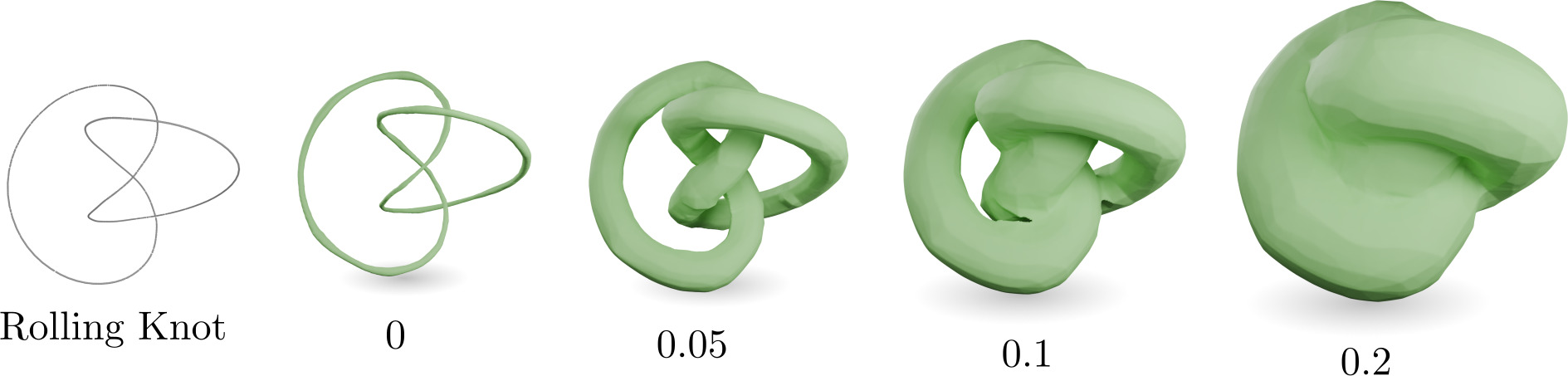} \\ ~ \\
\includegraphics[width=\textwidth]{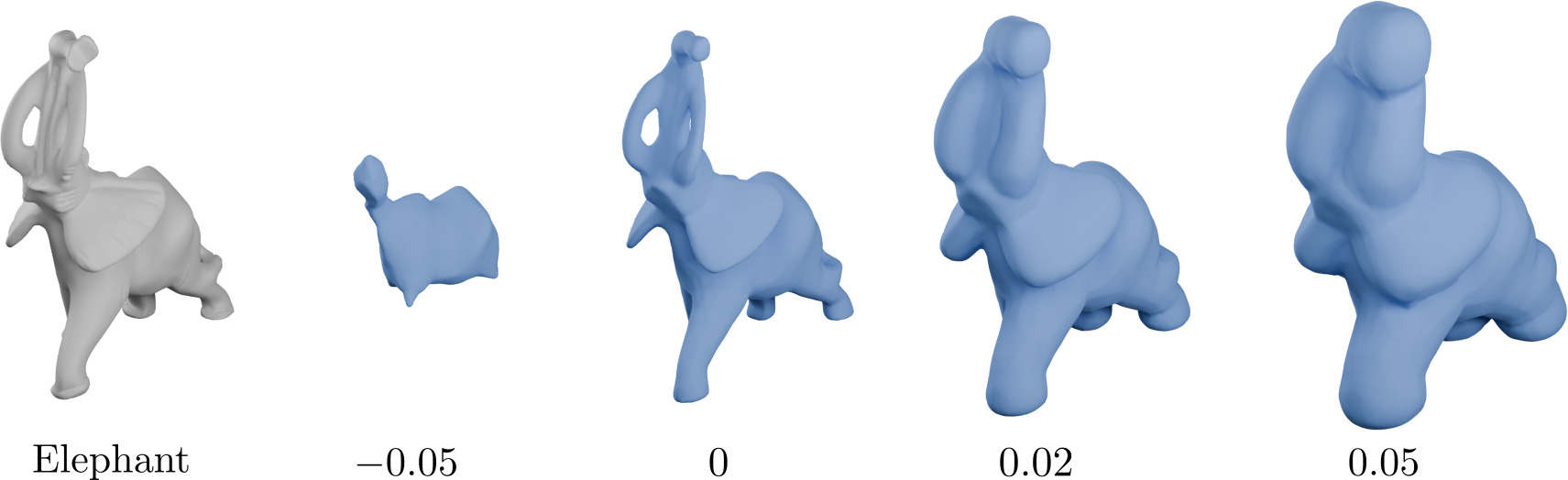} \\ ~ \\
\includegraphics[width=\textwidth]{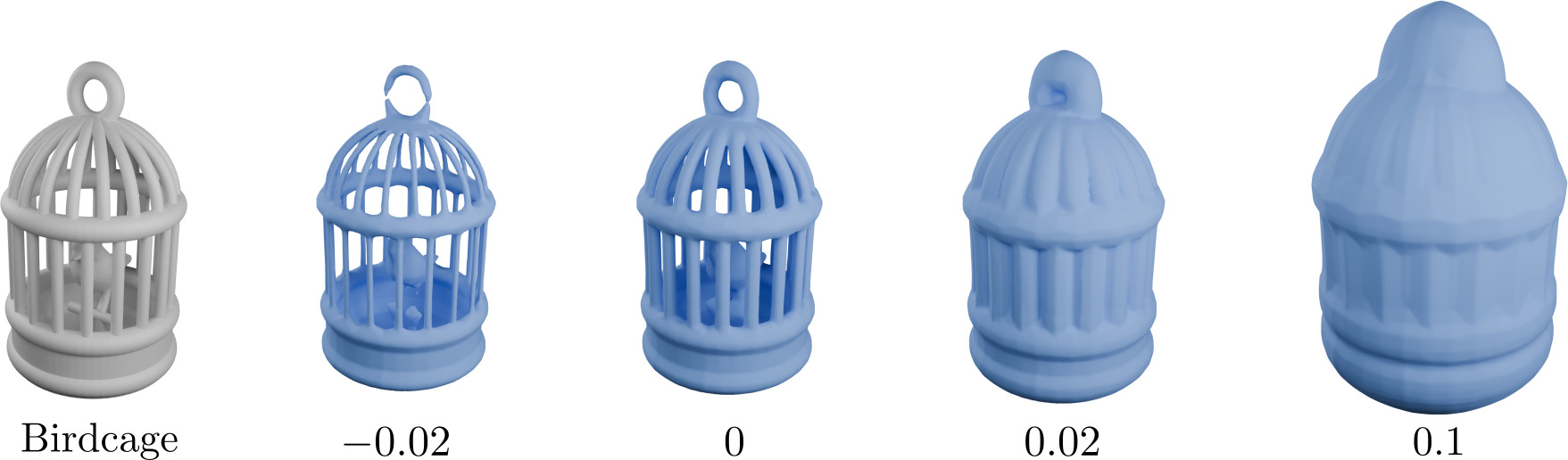} \\ ~ \\
\includegraphics[width=\textwidth]{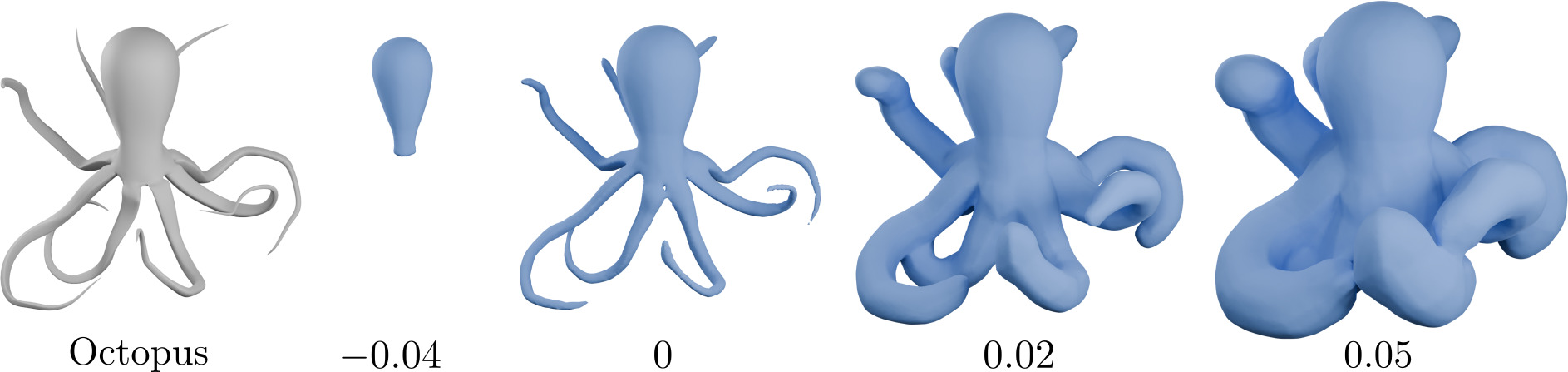} \\ ~ \\
\includegraphics[width=\textwidth]{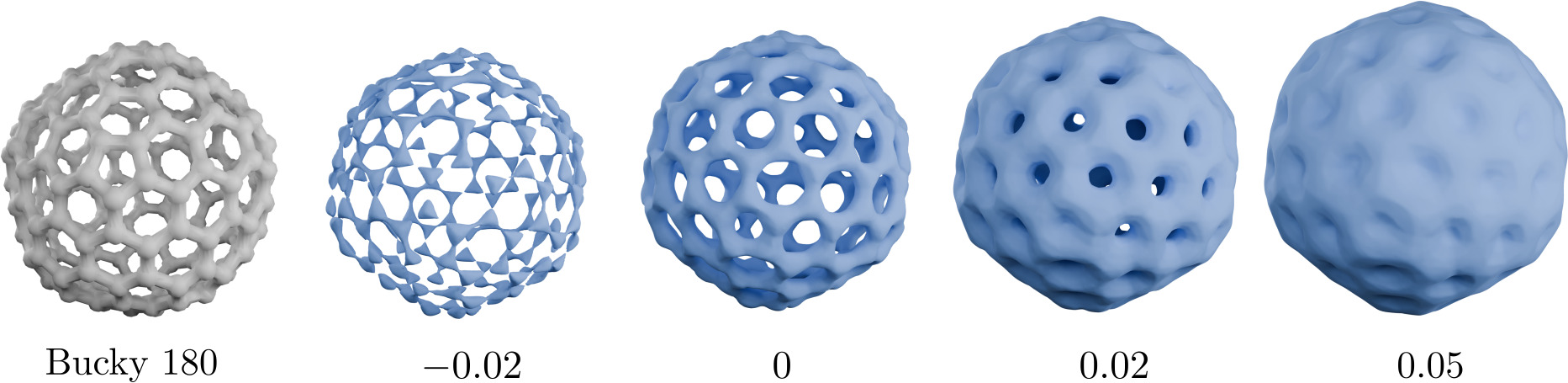} \\ ~ \\
\includegraphics[width=\textwidth]{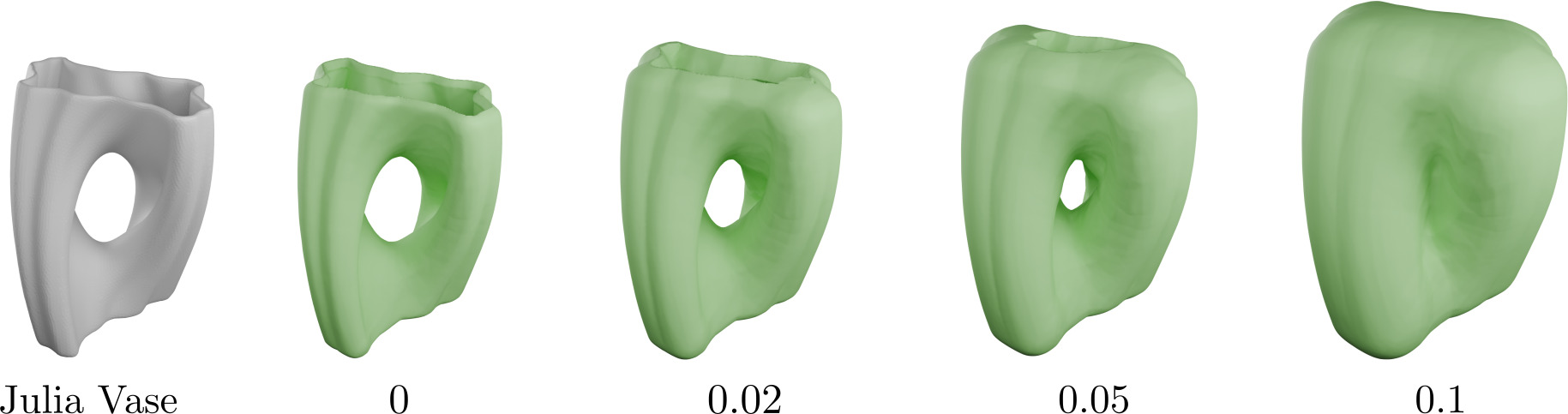} \\ ~ \\
\includegraphics[width=\textwidth]{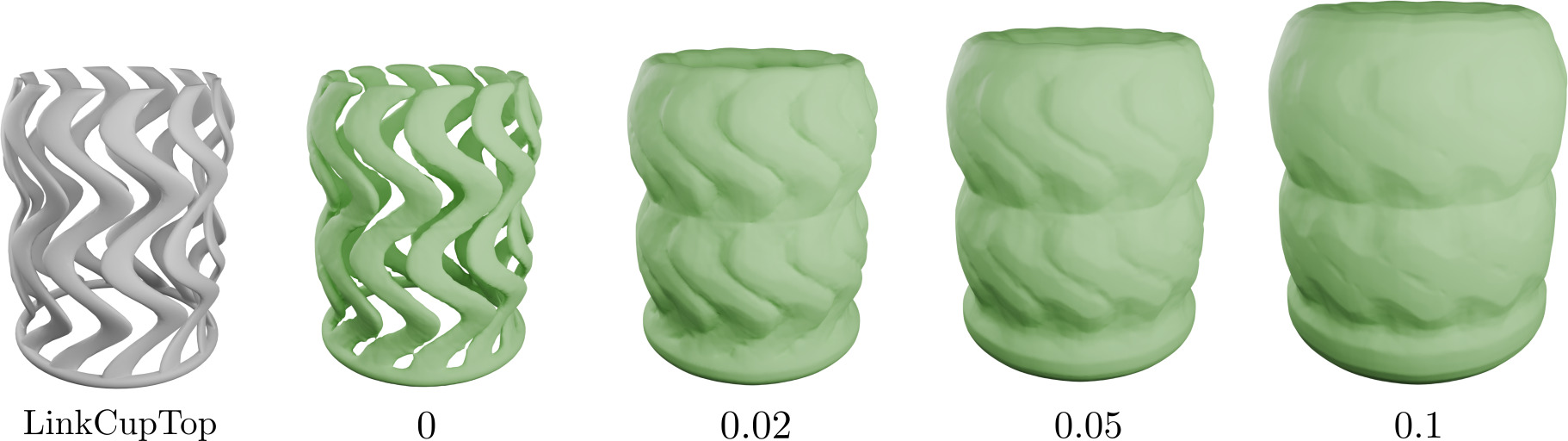} \\ ~ \\
\includegraphics[width=\textwidth]{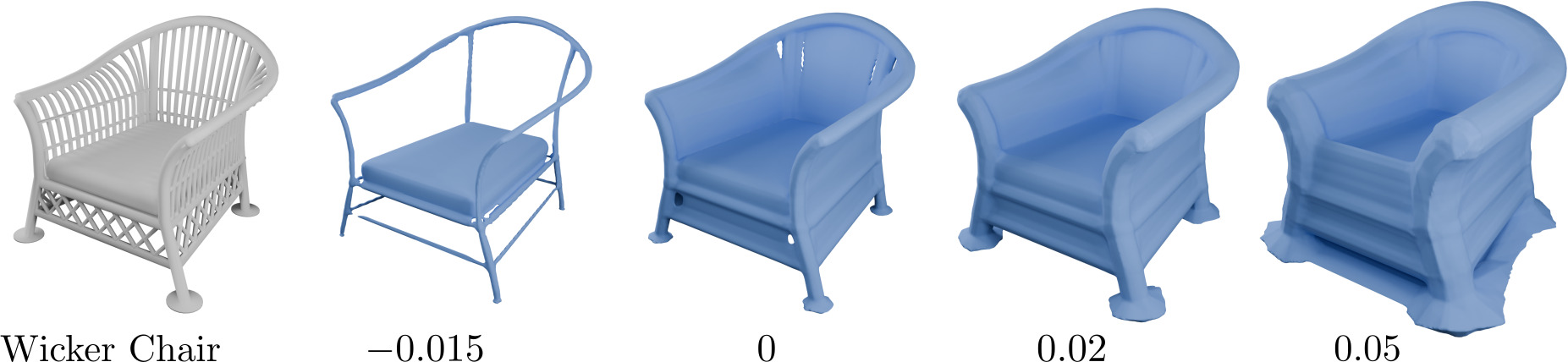} \\ ~ \\
\includegraphics[width=\textwidth]{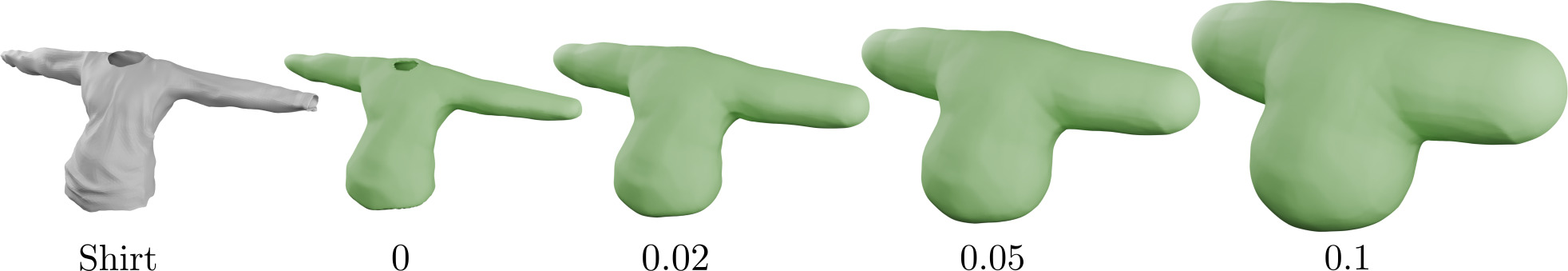} \\ ~ \\
\includegraphics[width=\textwidth]{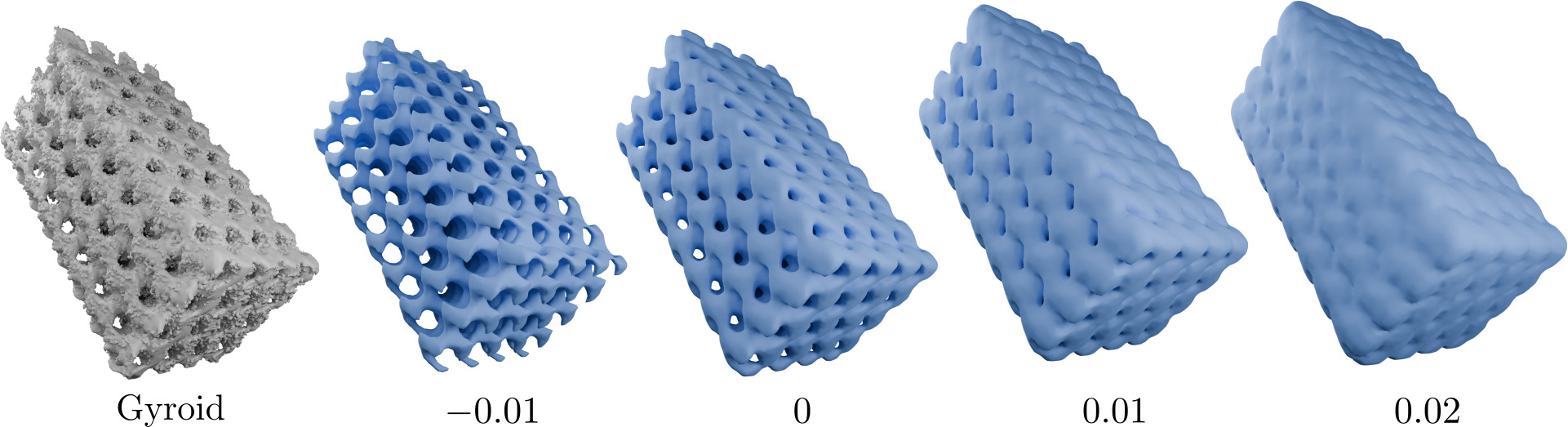} \\
\newpage

\section{Sampling and Ray marching}

Here, we perform geometrical queries on different models, namely sampling of the zero level set (left column, green points), sampling of the medial axis (middle column, red points) and ray marching rendering (right column).

\begin{center}
\includegraphics[width=0.9\textwidth]{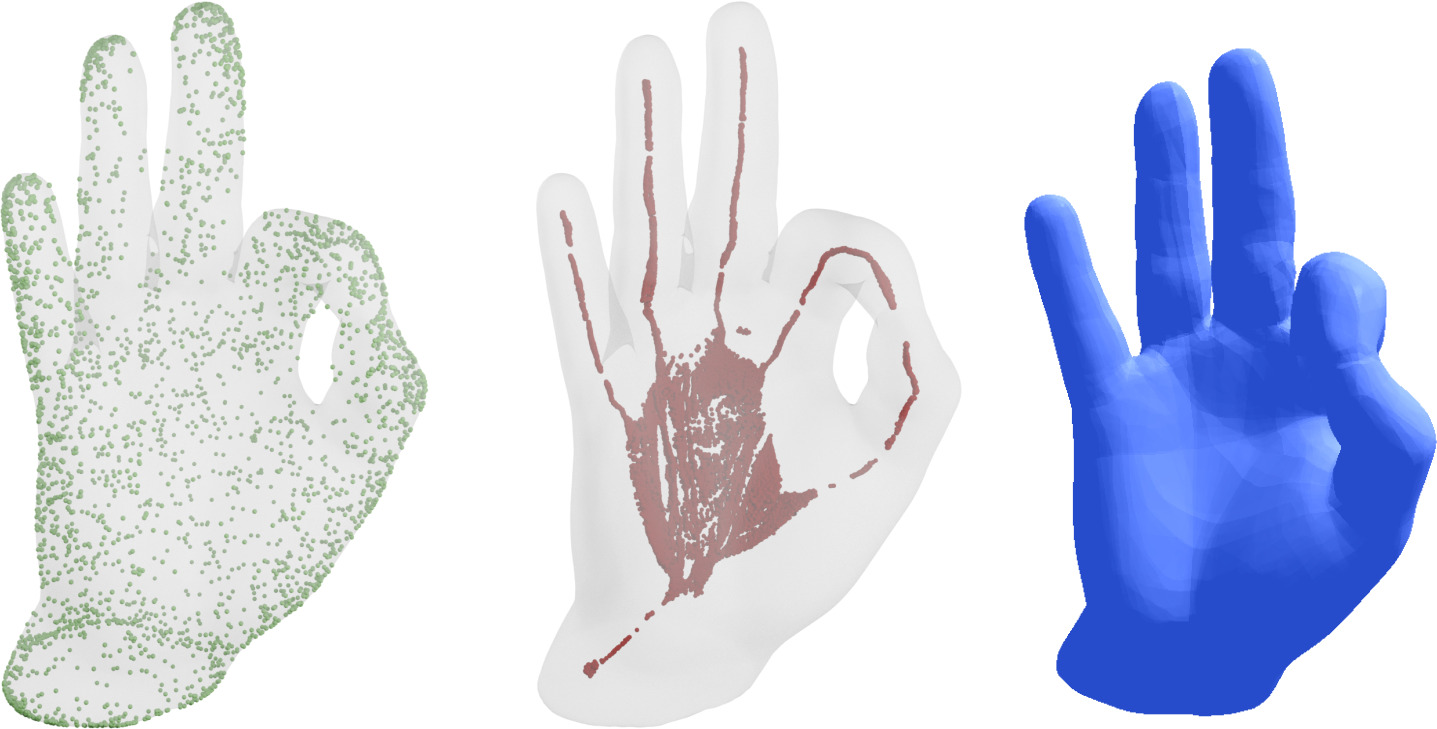} \\ \large Hand \vspace{1cm} \\
\includegraphics[width=0.9\textwidth]{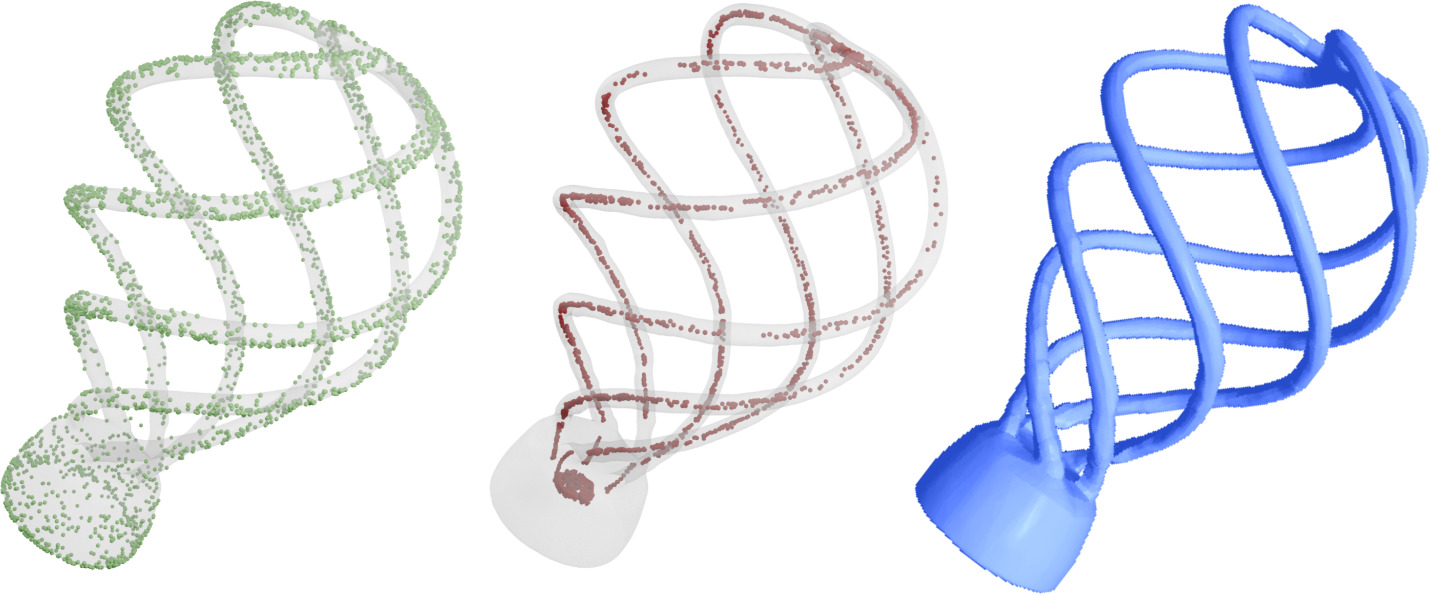} \\ \large Lightbulb \vspace{1cm} \\
\includegraphics[width=0.9\textwidth]{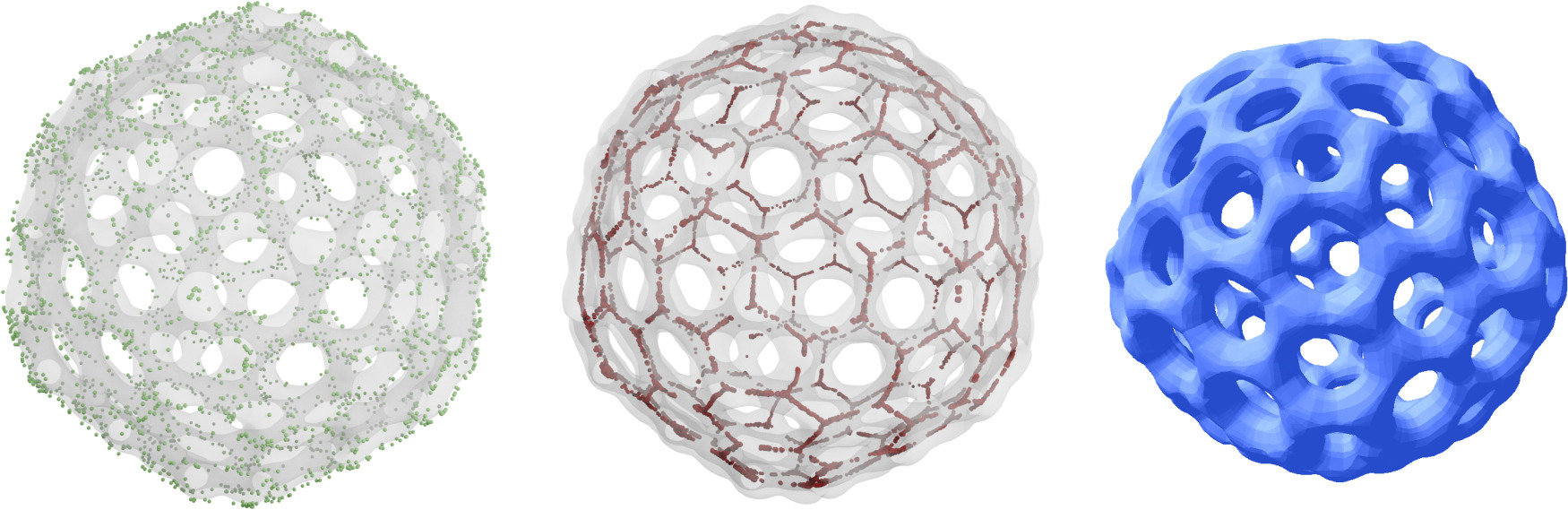} \\ \large Bucky180 \vspace{1cm} \\
\includegraphics[width=0.9\textwidth]{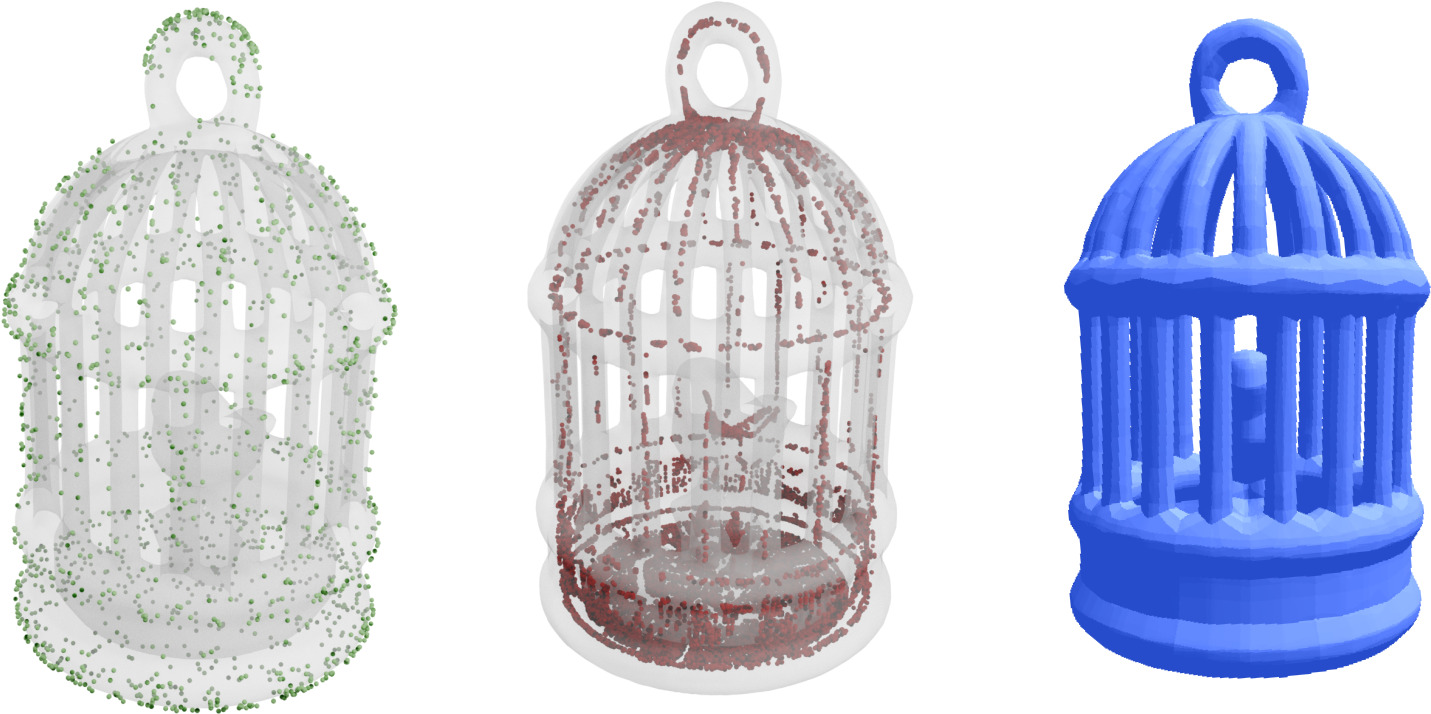} \\ \large Birdcage \vspace{1cm} \\
\end{center}



\bibliographystyle{eg-alpha-doi} 
\bibliography{biblio_zotero, biblio_other}